%% file: arxiv.tex
\documentclass[journal]{IEEEtran}

\ifCLASSINFOpdf

\else

\fi

\usepackage{graphicx}
\usepackage{times}
\usepackage{epsfig}
\usepackage{graphicx}
\usepackage{amsmath}
\usepackage{amssymb}
\usepackage{pifont}
\usepackage{float}
\usepackage{xcolor}
\usepackage{bm}
\usepackage{times}
\usepackage{epsfig}
\usepackage{graphicx}
\usepackage{float}
\floatstyle{plaintop}
\restylefloat{table}
\usepackage{multirow}
\newcommand{\xmark}{\ding{55}}%
\newcommand{\cmark}{\ding{51}}%
\usepackage{array}
\newcolumntype{x}[1]{>{\centering\arraybackslash\hspace{0pt}}p{#1}}
\usepackage{subcaption}
\usepackage[symbol]{footmisc}
\usepackage{wrapfig}
\usepackage{booktabs}
\usepackage{stfloats, caption}%

\usepackage{times}
\usepackage{epsfig}
\usepackage{graphicx}
\usepackage{amsmath}
\usepackage{amssymb}
\usepackage{footnote}
\usepackage{dsfont}

\usepackage{comment}
\usepackage{blindtext}
\usepackage{morewrites}
\usepackage{makecell}
\usepackage{pifont}%
\usepackage{multirow}
\usepackage{soul}

\begin{document}

\title{Vec2Face-v2: Unveil Human Faces from their Blackbox Features via Attention-based Network in  Face Recognition}

\author{Thanh-Dat~Truong,~\IEEEmembership{Student Member,~IEEE,}
        Chi~Nhan~Duong,~\IEEEmembership{Member,~IEEE,}
        Ngan~Le,~\IEEEmembership{Member,~IEEE,}
        Marios~Savvides,~\IEEEmembership{Senior~Member,~IEEE,}%
        ~and~Khoa~Luu$^{*}$,~\IEEEmembership{Member,~IEEE,}
\thanks{Thanh-Dat Truong, Ngan Le, and Khoa Luu are with the Computer Vision and Image Understanding Lab, University of Arkansas, USA}%
\thanks{Chi Nhan Duong is with the Department of Computer Science and Software Engineering, Concordia University, Canada}%
\thanks{Marios~Savvide is Carnegie Mellon University, USA}
\thanks{$^{*}$Corresponding Author: Khoa Luu, email: \texttt{khoaluu@uark.edu}}

\markboth{IEEE TRANSACTIONS ON IMAGE PROCESSING}%
{Truong \MakeLowercase{\textit{et al.}}: Vec2Face-v2: Unveil Human Faces from their Blackbox Features via Attention-based Network in  Face Recognition}
}

\maketitle

\begin{abstract}
In this work, we investigate the problem of face reconstruction given a facial feature representation extracted from a blackbox face recognition engine. Indeed, it is a very challenging problem in practice due to the limitations of abstracted information from the engine. We, therefore, introduce a new method named Attention-based Bijective Generative Adversarial Networks in a Distillation framework (DAB-GAN) to synthesize the faces of a subject given his/her extracted face recognition features.  Given any unconstrained unseen facial features of a subject, the DAB-GAN can reconstruct his/her facial images in high definition. The DAB-GAN method includes a novel attention-based generative structure with the newly defined Bijective Metrics Learning approach. The framework starts by introducing a bijective metric so that the distance measurement and metric learning process can be directly adopted in the image domain for an image reconstruction task. The information from the blackbox face recognition engine will be optimally exploited using the global distillation process. Then an attention-based generator is presented for a highly robust generator to synthesize realistic faces with ID preservation. We have evaluated our method on the challenging face recognition databases, i.e., CelebA, LFW, CFP-FP, CP-LFW, AgeDB, CA-LFW, and consistently achieved state-of-the-art results. The advancement of DAB-GAN is also proven in both image realism and ID preservation properties.
\end{abstract}


\section{Introduction}\label{sec:introduction}

\begin{figure*}[t]
	\centering \includegraphics[width=1.0\textwidth]{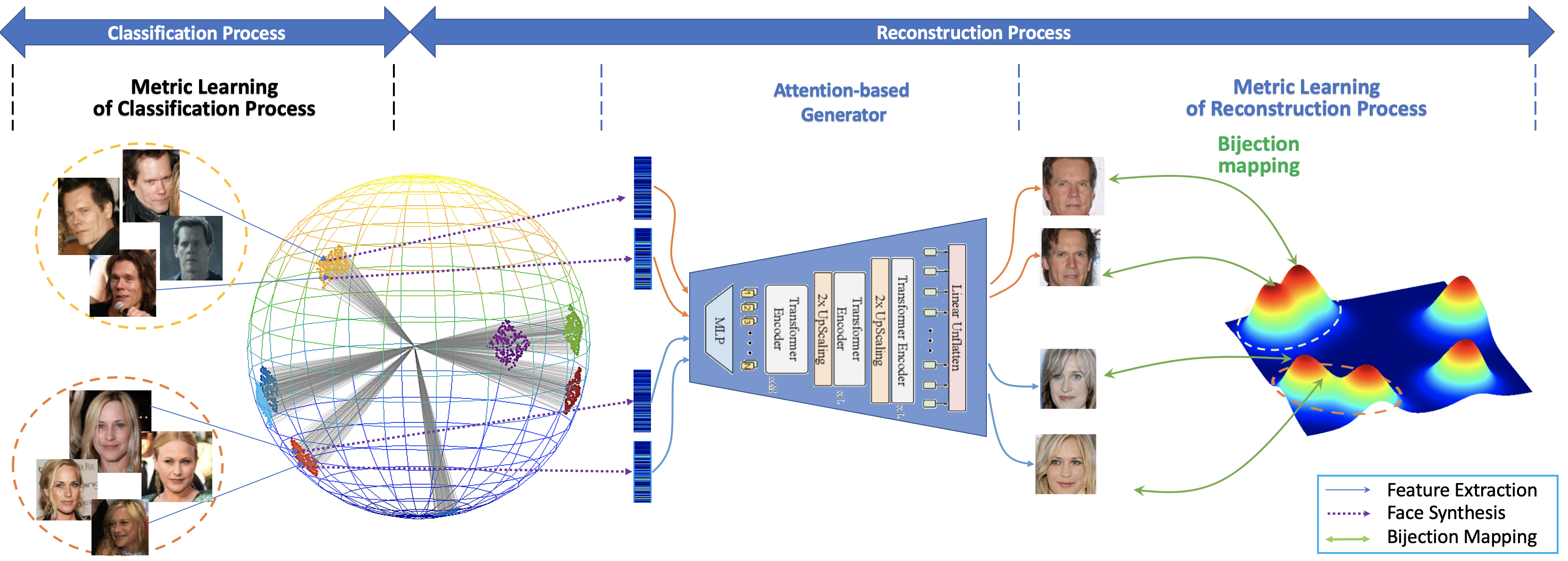}
	\small
	\caption{\textbf{Metric Learning for Image Reconstruction.} By maintaining the one-to-one mapping via a bijection, the distance between images can be directly and intuitively measured and enhances the metric learning process for image reconstruction. 
 } %
	\label{fig:BijectiveMetricLearningIllutration}
\end{figure*}

Face recognition \cite{zhao2003face} has recently matured and achieved remarkable accuracy against a large number of classes. %
There have been many approaches proposed in the literature that can be divided into two groups: i.e., \textit{``closed-set''} and \textit{`open-set`} image recognition. While the first group aims at recognizing only classes that are seen from training data, the second group relaxes the constraint on predefined classes and provides more power to recognize classes that are not included in its training data.
A typical \textit{open-set} recognition system is designed in two main phases, i.e., feature extraction and feature comparison. The role of feature extraction is more important since it directly determines the robustness of the engine. This operator defines an embedding process mapping input facial images into a higher-level latent space where embedded features extracted from photos of the same subject distribute within a small margin.
Moreover, to protect the technologies, most recognition systems are usually set into a \textit{blackbox} mode before their deployment stage. Therefore, in terms of both ``\textit{blackbox}'' and ``\textit{open-set}'' criteria, there are limited apparent attempts to inverse that embedding process for reconstructing input images of a subject given the extracted features from those \textit{blackbox} recognition engines. 
We refer this as a \textit{``feature reconstruction''} problem.

Some Adversarial Attack approaches \cite{pmlr-v80-ilyas18a,ilyas2018prior,Thys_2019_CVPR_Workshops,bousmalis2017unsupervised,tzeng2017adversarial} have partially addressed this task in the form of analyzing the gradients of recognition engine's outputs to generate adversarial examples that mislead the recognition engines' behaviors.
Either in the context of  whitebox or blackbox modes, these approaches are usually limited their working domain of predefined classes from the training set of the recognition engines. More importantly, the ultimate goal of these approaches is to generate imperceptible perturbations (i.e., random patterns) to be added to given input signals rather than synthesizing new images from scratch. 
Other methods
\cite{pmlr-v80-athalye18b, Cole_2017_CVPR,Dosovitskiy_2016_CVPR, NIPS2018_8052,Zhmoginov2016InvertingFE}
to reverse the process of feature extraction from classification models
and to explain the actual ``knowledge'' embedded in these models are also introduced in the literature. Although they can directly produce images without prior inputs as previous approaches, their crucial assumption is that the classifier structure is accessible, i.e., whitebox setting. 
Meanwhile, our goal focuses on a more challenging reconstruction task with a \textit{blackbox} recognition model. Firstly, this process \textit{reconstructs faces from scratch} without any hint from input images. %
Secondly, in a blackbox setting, there is \textit{no information about the engine's structure}, and, therefore, it is unable to directly exploit knowledge from the inverse mapping process (i.e., back-propagation). 
Thirdly, the embedded features from a recognition engine are for \textit{open-set problem} where no label information is available. More importantly, the subjects or classes to be reconstructed may have never been seen during the training process of the recognition engine.

\begin{table*}[b]
	\centering
    \setlength{\tabcolsep}{3pt}
	\caption{Comparisons of our DibiGAN and other unrestricted synthesis methods. Image Reconstruction (Img\_Recon), Feature Representation (Feat), Guided Image (Img$_G$), Feature Conditional (Feat\_Cond), Neighborly Deconvolution (NB\_Deconv), Optimization (Opt), Face Recovery (FaRe), Zero-order Optimization (ZO-Opt), Auto-Encoder (AE).} 
	\footnotesize
    \resizebox{\textwidth}{!}{  
	\begin{tabular}{| l | c c c c c c c|} %
            \toprule
		& \textbf{Vec2Face-V2} & GMI \cite{zhang2020secret} & FaRe \cite{razzhigaev2020black} & NBNet \cite{mai2018reconstruction}  & SynNormFace \cite{Cole_2017_CVPR}  & IFaceRec  \cite{Zhmoginov2016InvertingFE} & INVREP \cite{conf/cvpr/MahendranV15} \\
		
            \midrule

		\textbf{Input} & Feat & & & Feat  & Feat & Feat + Img$_G$ & Feat \\ %
		\textbf{Signal Type} & \textbf{Face} & Face & Face & Face  & Face & Face & Face \\ %
		\textbf{Feature Type} & CNN-based & & & CNN-based & CNN-based & CNN-based & CNN-based \\
            \midrule
		\begin{tabular}{@{}l@{}}\textbf{Generator}  \textbf{Structure} \end{tabular}   & \begin{tabular}{@{}c@{}}\textbf{Feat\_Cond} \end{tabular} & AE & \begin{tabular}{@{}l@{}}ZO-Opt \end{tabular} & \begin{tabular}{@{}c@{}}NB\_Deconv \end{tabular}  & \begin{tabular}{@{}c@{}} MLP + CNN \end{tabular}& DeConvNet & Opt \\
		
		\textbf{Blackbox Support} & \cmark & \xmark & \cmark &\cmark  & \xmark & \xmark & \xmark \\
            \midrule
		\begin{tabular}{@{}l@{}}\textbf{Img\_Recon Metric} \end{tabular}& \begin{tabular}{@{}l@{}} \textbf{Bijective} \end{tabular}& \xmark & \xmark & \xmark & \xmark & \xmark & \xmark \\ %
  \begin{tabular}{@{}l@{}}\textbf{Exploited Knowledge}\\\textbf{from Classifier} \end{tabular} & \begin{tabular}{@{}c@{}}\textbf{Fully}\\\textbf{(Distillation)} \end{tabular}& \begin{tabular}{@{}l@{}}Partially \end{tabular} & \begin{tabular}{@{}c@{}}Fully\\(Whitebox) \end{tabular}  & \begin{tabular}{@{}l@{}}Partially \end{tabular}& \begin{tabular}{@{}c@{}}Fully\\(Whitebox) \end{tabular}& \begin{tabular}{@{}c@{}}Fully\\(Whitebox) \end{tabular} & \begin{tabular}{@{}c@{}}Fully\\(Whitebox) \end{tabular} \\ 	

            \bottomrule

	\end{tabular}\label{tab:TenMethodSumm}
	}
\end{table*}

In the scope of this work, we assume that the recognition engines are primarily developed by deep Convolutional Neural Networks (CNN) that dominate recent state-of-the-art results. We also assume there is no further post-processing after the step of CNN feature extraction.
This work presents a novel generative structure, namely Attention-based Bijective Generative Adversarial Networks in a Distillation framework (DAB-GAN), with Bijective Metric Learning for the image reconstruction task. 
Far apart from previous approaches where the introduced metrics mainly fulfill the discriminative criteria of the classification process, the proposed Bijective Metric Learning with bijection (one-to-one mapping) property provides a more  effective, natural way to measure the distance between images for image reconstruction tasks. 
A preliminary version of our work can be found in \cite{duong2020vec2face}. In our prior work, the proposed DiBiGAN with Bijective Metric is able to synthesize faces with a wide range of in-the-wild variations against different face-matching engines and demonstrate the advantages of synthesizing realistic faces with the subject’s visual identity. 
In this paper, we further exploit the effect of the generator on the synthetic results. 
{
In particular, although the CNN-based generator has achieved remarkable performance, it still remains several limitations. In particular, the local receptive field and inflexible computation of the CNN network limit its ability to model long-range dependencies \cite{jiang2021transgan, hudson2021generative}. Thus, the model is not able to generalize the knowledge of the global shapes and structures and therefore, is limited to modeling complex patterns and structures. 
In addition, the optimization and stability of the CNN-based generator \cite{hudson2021generative} pose an ill-posed problem in facial reconstruction due to their fundamental difficulty in coordinating among details of facial characteristics across facial images.
Therefore, to address these limitations and further improve the robustness and reliability of our DAB-GAN, we propose an attention-based generator designed based on Transformers. As a result, the model represented in this work is more advanced in terms of the synthesizing capability of our attention-based generator.
}
{
In addition, as mentioned in our preliminary work, the direct reconstruction loss (e.g., $\ell_2$) remains limited in the problem of feature reconstruction. In this paper, to further illustrate the robustness of our \textit{Bijective Metric} against the feature reconstruction problem, we will mathematically prove the generalizability of our proposed Bijective Metric over the standard direct metrics in the image space based on the Lipschitz continuity property. 
}
In summary, the contributions of this work are six-fold.

\begin{itemize} %
    \item A novel parametric metric, i.e. \textbf{\textit{Bijective Metric}}, for the \textit{feature reconstruction} task is presented. By analyzing limitations of common metrics in feature reconstruction, i.e., non-parametric (e.g., $\ell_p$) and parametric metric (e.g., classifier-based metric),  we propose a novel \textit{Bijective Metric Learning} with bijection (one-to-one mapping) property so that the distance between latent representations in the latent space are equivalent to the distance those images. Therefore, our proposal provides a more effective and natural metric learning approach to the feature reconstruction problem. 

    \item We prove the generalizability of our \textit{Bijective Metric} over the reconstruction metric based by using the Lipschitz continuity. This further confirms the robustness and generalizability of our proposed \textit{Bijective Metric} loss to the feature reconstruction problem.  
    
    \item We introduce a novel \textbf{\textit{Attention-based Generator}} that is purely designed based on the Transformer architecture. The novel attention-based generator is able to capture global and local information and model complex visual structures.
    
    \item \textit{\textbf{Different aspects of the distillation process}} for the image reconstruction task in a blackbox mode are also exploited. They include distilled knowledge from the blackbox image recognition and ID knowledge extracted from a real structure of the object or subject.

    \item A \textbf{\textit{Feature-conditional Generator Structure}} learned by \textbf{\textit{Exponential Weighting Decay Strategy}} within a Generative Adversarial Network (GAN)-based framework is introduced to improve the robustness of the generator to synthesize realistic images with identity preservation.
    
    \item The \textbf{\textit{intensive evaluations on face recognition benchmarks 
    }} have shown the performance of our proposal in both image realism and identity preservation.
\end{itemize}
To the best of our knowledge, this is one of the first metric learning methods for image reconstruction (Table \ref{tab:TenMethodSumm}).
The paper is organized as follows. Section \ref{sec:RelatedWork} reviews some recent approaches to synthesizing images from given (deep) features and Transformer backgrounds. 
Section \ref{sec:FeatReconDefine} presents the feature reconstruction problem w.r.t. blackbox image recognition engines and the proposed network architecture.
Then, we analyze the choices of available metrics being adopted and their limitations followed by proposing a novel bijective metric.
Then, different  aspects  of  the  distillation  process and the proposed Generator Structure of DAB-GAN for  the  image reconstruction task in  a  blackbox  mode  are  also  exploited and formulated.
Finally, the experimental results together with limitations and conclusions are given in Sections \ref{sec:Experiments}-\ref{sec:Conclusion}.

\section{Related Work} \label{sec:RelatedWork}

Face recognition has been well-studied for years. Many prior works have introduced well-designed network architectures and losses to improve the robustness and discriminativeness of deep facial features.
Later, synthesizing images from their deep features has brought several interests from the community.
According to the motivations and approaches to be used, they can be divided into two groups, i.e., unrestricted and adversarial synthesis methods. 

\subsection{Face Recognition}

Parkhi et al. \cite{Parkhi15} first introduced a VGG-based network for face recognition.
Later, many other network backbones have been successfully applied for face recognition, e.g., Residual Networks \cite{he2016deep}, Mobile Networks \cite{howard2017mobilenets, sandler2018mobilenetv2}, Vision Transformers \cite{zhong2021face, sun2022part}.
To improve the discriminativeness in deep feature representations, several works have proposed better learning objectives.
Hoffer et al. \cite{hoffer2015deep} first presented the Triplet loss that enforces the features of the same class being closer while pushing features of different classes far away based on the triplet data. However, mining meaningful triplets remain challenging in the Triplet loss. Then, several works improved this step by introducing effective sampling strategies \cite{rippel2015metric, oh2016deep, schroff2015facenet, sohn2016improved}. Margin-based Softmax methods \cite{deng2019arcface, wang2018cosface, liu2017sphereface} have been widely used recently. These approaches focus on  incorporating margin penalty into the Softmax loss.
Compared to prior metric learning methods (e.g., Triplet \cite{hoffer2015deep, triplet_reid}), margin-based Softmax methods are memory efficient while still maintaining the discriminative representations. Several subsequent works further improved the margin-based Softmax loss by presenting adaptive parameters \cite{kim2022adaface, zhang2019adacos}, or inter-class discrepancy alignment \cite{liu2021inter}.
Since many facial datasets contain ambiguous and inaccurate labels \cite{deng2019arcface, yi2014learning}, learning face recognition models with massive noisy data remains a challenging problem. Prior works improve the robustness of the face recognition models against noisy labels by using relabeling \cite{wu2018light}, clustering loss \cite{nguyen2021clusformer, nguyen2023fairness}, or re-weighting methods \cite{wang2019co}. Face recognition with sub-classes \cite{zhu2004optimal, zhu2006subclass, wan2017separability, qian2019softtriple} has gained much attention recently. This approach depicts the underlying distribution of facial data, thus, further improves the performance of the face recognition models against various conditions of faces (e.g., front-view and side-view faces). 

\subsection{Unrestricted Synthesis Methods}

The approaches in this direction 
focus on reconstructing an image from scratch given its high-level representation. 
Since the mapping is from a low-dimensional latent space to a highly nonlinear image space, several regularizations have to be applied, e.g., Gaussian Blur \cite{yosinski2015understanding} for %
high-frequency samples or Total Variation \cite{conf/cvpr/MahendranV15} for maintaining piece-wise constant patches. %
These optimization-based techniques are  limited with high computation and unrealistic reconstructions. Later, Dosovitskiy et al. \cite{Dosovitskiy_2016_CVPR} proposed to reconstruct the image from its shallow (i.e. HOG, SIFT) and deep features using a Neural Network (NN).
Zhmoginov et al. \cite{Zhmoginov2016InvertingFE} presented an iterative method to invert Facenet \cite{schroff2015facenet} feature with feed-forward NN. %
Cole et al. \cite{Cole_2017_CVPR} proposed an autoencoder structure to map the features to the frontal neutral face of the subject. Yang et al. \cite{Yang:2019:NNI:3319535.3354261} also adopted an autoencoder for the model inversion task.
Generally, to produce better-synthesized quality, these approaches require full access to the deep structure 
to exploit the gradient information from the embedding process.
Mai et al. \cite{mai2018reconstruction} developed a neighborly deconvolutional network to support the blackbox mode. However, with only pixel and perceptual \cite{johnson2016perceptual} losses, there are limitations of ID preservation when synthesizing different features of the same subject. 
In this work, we further address and solve this issue using Bijective Metric Learning and Distillation Knowledge for the reconstruction task. 

\subsection{Adversarial Synthesis Methods}
Rather than reconstructing the images from scratch, Adversarial approaches aim at generating unnoticeable perturbations from input images for adversarial examples to mislead the behavior of a deep structure. %
Either directly accessing or indirectly approximating gradients, adversarial examples are created by maximize corresponding loss which can fool a classifier \cite{pmlr-v80-athalye18b, Brunner2019CopyAP, Cheng2019ImprovingBA, pmlr-v80-ilyas18a, ilyas2018prior,   Liu2016DelvingIT,MoosaviDezfooli2015DeepFoolAS, Shukla2019BlackboxAA, Thys_2019_CVPR_Workshops}. Ilyas et al. \cite{ilyas2018prior} proposed bandit optimization to exploit prior information about the gradient of the deep learning models.  Later, Ilyas et al. \cite{pmlr-v80-ilyas18a} introduced Natural Evolutionary Strategies to enable query-efficient generation of black-box adversarial examples.
Other knowledge from the blackbox classifier are also exploited for this task \cite{Thys_2019_CVPR_Workshops, pmlr-v80-athalye18b, NIPS2018_8052}.
Generally, although the approaches in this direction tried to extract the gradient information from a blackbox classifier, their goal is mainly to mislead the behaviors of the classifier with respect to a pre-defined set of classes. Therefore, they are closed-set approaches. Meanwhile, in our work, the proposed framework can reconstruct the faces of subjects that have not been seen in the training process of the classifier.

\subsection{Transformer for Image Synthesis}
Transformer is initially used for Natural Language Processing (NLP) tasks, where it has demonstrated considerable progress \cite{vaswani2017attention, devlin2018bert, brown2020language}. Vaswani et al. \cite{vaswani2017attention}, for example, was the first to propose a transformer based on an attention mechanism for machine translation and English constituency parsing tasks.
Devlin et al. \cite{devlin2018bert} proposed a novel language representation model called BERT (short for Bidirectional Encoder Representations from Transformers), which pre-trains a transformer on the unlabeled text while accounting for each word's context (it is bidirectional). BERT achieved state-of-the-art performance on NLP tasks. %
Brown et al.\cite{brown2020language} used 45 TB of compressed plaintext data to build a huge transformer-based model termed GPT-3 (short for Generative Pre-trained Transformer3) with 175 billion parameters. It delivered strong results on a variety of downstream natural language tasks without the need for fine-tuning. These transformer-based models, with their high representation capacity, have made substantial advances in NLP. Researchers have lately adapted Transformer to computer vision applications, inspired by the significant success of transformer designs in the field of NLP. CNNs are regarded as the core component in vision applications \cite{he2016deep, ren2015faster}. However, Transformer is currently showing promise as a viable replacement to CNN. Vision Transformer (ViT) is another Transformer model that applies a pure Transformer directly on picture patch sequences. Dosovitskiy et al. \cite{dosovitskiy2020image} first introduced ViT to image classification and has shown its state-of-the-art performance on various image recognition benchmarks. Transformer has been used to solve a wide range of different computer vision problems, including object recognition \cite{zhu2020deformable,carion2020end}, semantic segmentation \cite{xie2021segformer}, and video action recognition \cite{arnab2021vivit, truong2022direcformer, liu2022video}, Object Detection \cite{zhu2020deformable, carion2020end, zhang2022dino}. 

\noindent
\textbf{Transformer For Image Reconstruction}
Jiang et al. \cite{jiang2021transgan} introduced GAN architecture that consists of a generator based on Transformer to improve the resolution of features and a discriminator to obtain the low-level textures and semantic contexts. They also introduced a novel module of self-attention that improves the memory bottleneck to enhance high-resolution generation. Lin et al. \cite{lin2018st} introduced GAN architecture which utilizes Spatial Transformer Network (STN), which operates in the geometric warp parameter space. They proposed a sequential training strategy and an iterative STN warping scheme which proved better than the training of a single generator. Hudson et al. \cite{hudson2021generative} proposed GANformer using a bipartite structure that is able to model the long-range interactions and maintain the efficiency of the computation for high-resolution image synthesis. Then, it is further improved for the scene reconstruction task by equipping a strong explicit structure to capture context dependencies and interactions among objects in an image \cite{arad2021compositional}. Esser et al. \cite{esser2021taming} proved an efficient way of combining CNNs with Transformer to synthesize high-resolution images. It uses the CNNS to learn the context of the images and uses the Transformers to efficiently model the composition in high-resolution images. Chen et al. \cite{chen2020generative} trained a sequential transformer that predicts the pixels auto-regressively and this whole procedure is done without the knowledge of inputs. Recent works \cite{dubey2023transformerbased} have exploited the ViT \cite{dosovitskiy2020image} and Swin Transformer \cite{liu2021swin} for the reconstruction task due to their ability to preserve the local context.

\section{Feature Reconstruction Problem} \label{sec:FeatReconDefine}

In this section, we detail the feature reconstruction problem in face recognition. First, the formulation of this problem will be mathematically defined. Then, the proposed network architecture will be introduced using a CNN-based generator and an Attention-based generator. 

\subsection{Problem Formulation}

Given an input image $\mathbf{I}$ in the image space $\mathcal{I} \in \mathbb{R}^{H \times W \times 3}$ ($H, W$ are the height and width), a function $F: \mathcal{I} \to \mathcal{F}$ maps an input image $\mathbf{I} \in \mathcal{I}$ to its deep feature representation space $\mathcal{F} \in \mathbb{R}^{M}$ ($M$ is the feature dimension length). An additional function $C: \mathcal{F} \to \mathcal{Y}$ takes the deep feature $F(\mathbf{I})$ as input and produces the corresponding identity prediction of the subject in the space $\mathcal{Y} \in \mathbb{R}^{N}$ where $N$ is the number of predefined subject classes in the training dataset. Two types of reconstruction problems can be defined as follows.

\noindent
\textbf{Definition 1.} \textit{\textbf{Model Inversion:} Given blackbox functions $F$ and $C$; and a prediction identity vector $\mathbf{s}$ extracted from an unknown image $I$, i.e. $\mathbf{s} = [F \circ C] (\mathbf{I})$, the model inversion task will recover an input image ${\mathbf{I}}$ from $\mathbf{s}$ such that}
\begin{equation}
    \tilde{\mathbf{I}}^* = \arg \min_{\tilde{\mathbf{I}}} \mathcal{L}([F \circ C] (\tilde{\mathbf{I}}), \mathbf{s})
\end{equation}
\textit{where $\mathcal{L}$ defines some types of distance metrics.}

The approaches to the model inversion problem can be addressed by exploiting the relation between the input image and its class label. Additionally, the number of classes $N$ is predefined by the number of training subject identities, and thus, the model inversion is limited to images of the training sets. The problem, in this case, is also known as the \textbf{\textit{closed-set reconstruction}}.

\noindent
\textbf{Definition 2.} \textit{\textbf{Feature Reconstruction:}} Given a blackbox function $F$; and its embedding feature $\mathbf{f} = F(\mathbf{I})$ of an unknown image $\mathbf{I}$, feature reconstruction is to recover $\mathbf{I}$ from $\mathbf{f}$ by optimizing Eqn. \eqref{eqn:argI} as follows,

\begin{equation} \label{eqn:argI}
    \tilde{\mathbf{I}}^* = \arg \min_{\tilde{\mathbf{I}}} \mathcal{L}(F (\tilde{\mathbf{I}}), \mathbf{f}).
\end{equation}

Different from the first type, this problem relaxes the constraint on the number of classes, making it an \textbf{\textit{open-set reconstruction}} problem. This brings more challenging factors to the reconstruction process as the target reconstructed faces are not necessary to be seen by $F$ during its training process. Moreover, as parameters of $F$ is inaccessible, gradient-based approaches %
\cite{upchurch2017deep}
have become infeasible. A solution for this task is to  learn a reverse mapping function of $F$, i.e., \textit{generator}, $G: \mathcal{F} \to \mathcal{I}$ as in Eqn. \eqref{eqn:GeneratorFormulation}.
\begin{equation} \label{eqn:GeneratorFormulation}
\begin{split}
    \tilde{\mathbf{I}}^* &= G(\mathbf{f}; \theta_g)\\
    \theta_g &= \arg \min_{\theta} \mathbb{E}_{\mathbf{x} \sim p_I(\mathbf{x})} \left[ \mathcal{L}_G^x \left([G \circ F](\mathbf{x}; \theta), \mathbf{x}\right)\right]\\
     &= \arg \min_{\theta} \int
    \mathcal{L}_G^x \left(\tilde{\mathbf{x}}, \mathbf{x}\right) p_I(\mathbf{x}) d\mathbf{x}\\
\end{split}
\end{equation} 
where $\theta_g$ denotes the parameters of $G$, and $p_I(\mathbf{x})$ is the probability density function of $\mathbf{x}$.
In particular, the generator $G$ takes the deep feature in the deep representation space extracted by the network $F$ as input and produces the corresponding image in the image space where the reconstructed images $G(F(\mathbf{x}))$ are guarantee to be closed to its actual image $\mathbf{x}$ measured by a distance metric $\mathcal{L}_G^x`$.

Up to this point, there are two crucial problems related to the feature reconstruction problem, i.e., the design of the generator $G$ and the choice of distance $\mathcal{L}_G^x$. In Section \ref{subsec:proposedNetwork}, we further discuss the design of our proposed generators. Then, we analyze the optional selection of $\mathcal{L}_G^x$ followed by proposing a novel bijective metric.

\subsection{The Proposed Network Architecture} \label{subsec:proposedNetwork}

As discussed in the previous section, the design of generator $G$ plays an important role since it mainly affects the reconstructed results. 
It is straightforward to adopt the common generator structure \cite{karras2017progressive} for $G$ where the generator takes the $F(\mathbf{x})$ as an input. 
However, the native structure design limits the generality of the generator $G$. 
In particular, the common design of $G$ is a deterministic network where one input will produce only one image. 
In addition, along with the identity information, the input $F(\mathbf{x})$ may include other ``background" information, e.g., pose, illuminations, and expressions.
Therefore, if we only take $F(\mathbf{x})$ as an input, it means that we implicitly enforce $G$ to strictly model these factors. As a result, the reconstructed images will not be diversified and the training process of $G$ will be less efficient.

\subsubsection{CNN-based Generator with Feature-Condition Structure} \label{sec:Generator}

\begin{figure*}[t]
	\centering \includegraphics[width=1.0\textwidth]{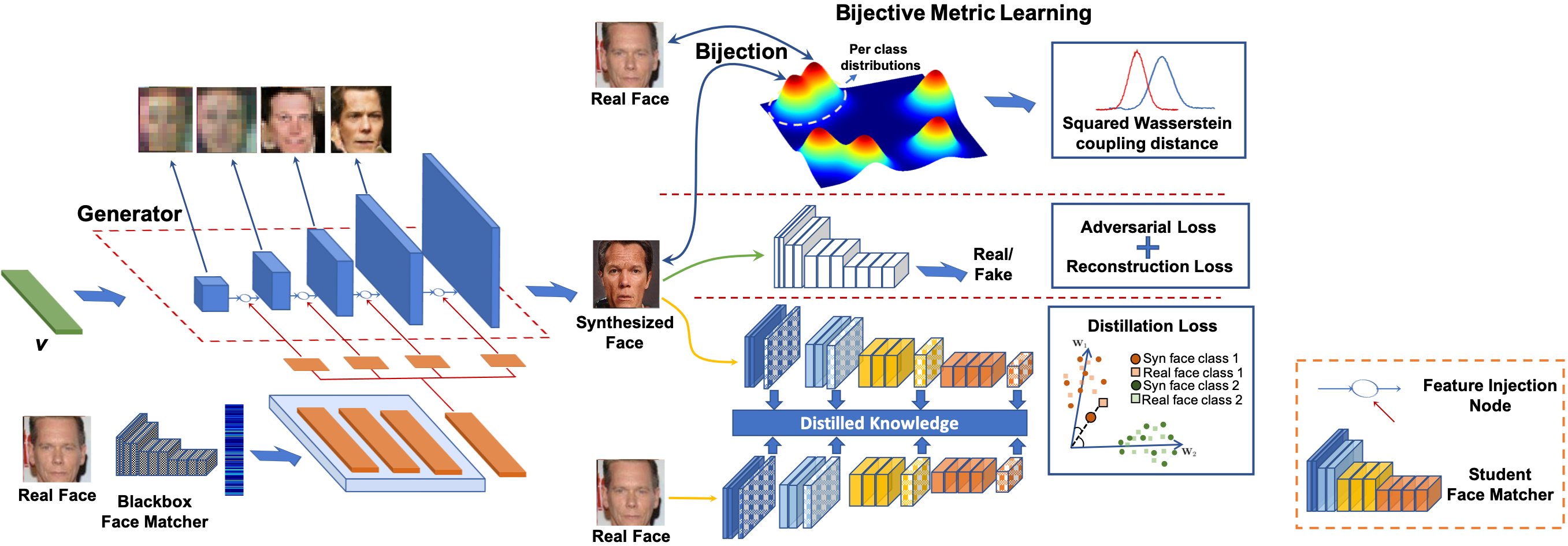}
	\caption{\textbf{The Proposed DAB-GAN Framework.} Given a high-level embedding representation,
    an \textit{Attention-based Generator}
	injects that representation throughout its structure as the conditional information for all scales. The cost functions are designed with \textit{Bijective Metric} to directly exploit ID distributions in the image domain, and \textit{Distillation Loss} to maximize the knowledge that could be extracted from the blackbox matcher. 
 }%
	\label{fig:ProposedFramework}
\end{figure*}

To address the aforementioned constraints, a novel feature-conditional structure is introduced, as shown in Fig. \ref{fig:ProposedFramework}. In this framework, a random variable $\mathbf{v}$ is adopted as an additional input represented for the background factors. With our design, the vector $\mathbf{v}$ will be a direct input of the generator, and the information of deep feature $F(\mathbf{x})$ will be progressively injected throughout the structure. By this design, the generator will gradually grow up the level of details of images and the feature $F(\mathbf{x})$ plays a role as the condition identity-related information for all reconstruction scales. Hence, the generator will give better synthetic results and the generality of the model is further enhanced. The detail of the generator will be described in the experiments in Section \ref{sec:Experiments}.

Although the CNN-based generator has achieved remarkable results in the image reconstruction tasks, this still suffers several issues. First, as the convolutional operators have the local receptive field, the CNN-based generator is not able to model long-range dependencies. 
Although we could design a CNN network with a large number of convolutional layers, it is inefficient since it could lose the feature resolution and fine-grained features. Also, optimizing a huge CNN network is a difficult problem. 
Second, the spatial invariance property of the convolutional layers leaves an issue on the adaptive ability of the generator to spatially varying and complex visual patterns. Moreover, even though we could design sophisticated CNN-based generators, training CNN-based generators within the GAN framework remains unstable and could result in mode collapse.
Therefore, despite the success of CNN-based generators, it is essential to design a generator that can address the mentioned problems of CNN-based networks. 

\subsubsection{Attention-based Generator with Transformer}

Motivated by the success of the Transformer in NLP and other vision tasks, the Transformer network offers a 
highly-adaptive architecture that is able to capture long-range context dependencies and dynamic interactions among image patches. Therefore, to address the problem of a CNN-based generator, we introduce a novel effective generalization of an attention-based generator for the feature reconstruction problem.
In particular, we design a convolution-free generator in which each block of the generator is purely designed by the Transformer block.
The condition identity-related information, e.g., $F(\mathbf{x})$, is progressively injected into all the network scales throughout the structure by the self-attention mechanism.
With the design of the Transformer, the generator can capture both local and global information through the attention mechanism.
Also, the transformer-based generator is able to capture long-range dependencies without passing numerous layers as a CNN-based generator.
As a result, the attention-based generator can model complex visual structures.
However, training a pure transformer-based generator is difficult since the vanilla Transformer network requires a large amount of memory when the image size scales up. 
To alleviate this issue, we adopt the memory-friendly transformer-based design of \cite{jiang2021transgan} to enhance memory efficiency and maintain the synthesizing capability.
With the design of our attention-based generator, our approach is able to address the issue of computational costs and the fundamental stability issues caused by the dense and excessive pairwise connectivity of the pure transformer [CITATIONS], and therefore, produce better facial reconstruction results.

\section{The Proposed Bijective Metric}

\subsection{The choices of $\mathcal{L}_G^x$ and Their Limitations}
\label{sec:LimitationClassifierMetric}

The quality of reconstruction results is primarily dependent on the choice of $\mathcal{L}_G^x$.
Several prior works \cite{smith2015bijective} have exploited different choices of $\mathcal{L}_G^x$ to produce good quality results. 
Overall, these choices could be divided into two groups: (1) \textit{non-parametric}, and (2) \textit{parametric} measurements.

\textbf{\textit{Non-parametric metrics}} (i.e., direct metrics) such as $\ell_p$ or Total Variation typically measure the pixel-level difference between the reconstructed and ground-truth images. Although these metrics are straightforward to be adopted and optimized, they are \textit{unstructured measurements} where a pixel is penalized independently to other pixels as well as the structured characteristics of the image. Moreover, they suffer other limitations in image quality. For example,  images synthesized with $\ell_2$ loss are surprisingly poor observed by humans \cite{6467150}. 

\textbf{\textit{Parametric metrics}} (i.e., indirect metrics or classifier-based metric) such as Adversarial loss or  Perceptual Loss have proved their advantages in several tasks, e.g., Image Style Transfer \cite{gatys2016image}, Image Synthesize \cite{reed2016generative}.
Typical parametric metrics are computed via additional mapping functions which are chosen according to 
the problem and the requirements of reconstruction quality. 
For example, the perceptual loss measures the differences between latent representations of images extracted via  a classifier pre-trained on the ImageNet. Therefore, this metric is able to capture more semantic information such as differences in object structures in its measurement. 
On the other hand, Adversarial Loss learns a binary classifier (i.e., a discriminator) to penalize the realism of the image. This type of metric, therefore, provides better reliable measurements in terms of image realism and semantic information preservation in comparison to non-parametric metrics.

Although these parametric metrics have shown their advantages in several tasks, there are limitations when only the blackbox function $F$ and its embedded features are given.
In particular, 
as shown in prior adversarial attack studies \cite{engstrom2019learning, santurkar2019computer}, as \textit{\textbf{the function $F$ is not an one-to-one mapping function from $\mathcal{I}$ to $\mathcal{F}$}}, it is straightforward to find \textit{two images sharing similar deep representations extracted by $F$ that are drastically different in the image content}. As a result, without prior information about the subject identity of an image, reconstructing it from scratch may fail in the case where the reconstructed image $\tilde{\mathbf{I}}$ is completely different from $\mathbf{I}$ but has similar deep feature representations.
The current classifier-based metrics, e.g., adversarial training losses or perceptual losses, are limited in maintaining the constraint ``\textit{the reconstructions of features of the same subject identity should be similar}''. This issue makes up the first criterion in our problem, i.e., \textbf{\textit{Identity Preservation}}.

Additionally, learning with parametric metrics requires the gradients of the network structures during the optimization process. However, 
since access to the network structure and intermediate features of the function $F$ is unavailable due to the constraint of the blackbox setting,
the function $G$ is unable to directly exploit valuable information from the gradients of $F$ and the intermediate representations. 
Therefore, the distance metrics defined via $F$, i.e., perceptual distance, are less effective compared to the whitebox setting. This points to the second criterion in our problem, i.e., \textbf{\textit{Blackbox Setting}}.

To tackle these constraints, we propose a novel bijective metric
to address these limitations for the feature reconstruction task. 
Moreover, a distillation process is also proposed to exploit further information from the given blackbox recognition engine for the reconstruction process.

\noindent

\noindent

\subsection{The Proposed Bijective Metric} \label{sec:ProposedMetric}

In this section, we initially present our proposed bijective metric for feature reconstruction that addresses the limitations of identity preservation.
Next, we discuss the learning procedure of our proposed bijective metrics.
Finally, we exploit the different aspects of the distillation approach to address the limitations of blackbox setting.

Many proposed metrics with sophisticated designs used for face recognition \cite{deng2019arcface, LiuNIPS18,liu2017sphereface,  wang2018cosface,wen2016discriminative, zhang2019adacos} have been employed to improve the discriminative capability of features by maximizing intra-class compactness and inter-class separability with a large margin.
Nevertheless, adopting these metrics for the feature reconstruction task is an infeasible solution since these metrics are limited as aforementioned in the previous section (the problem of \textbf{\textit{Identity Preservation}}).

Taking these limitations into consideration, we propose a bijective metric for the feature reconstruction task in which the mapping function from the image space to the deep latent space is one-to-one. 
Thanks to the bijective property, the distance between images will be equivalent to the distance between their deep latent features. 
As a result, the proposed metric is more aligned with the image domain and can be efficiently adopted for the reconstruction task. 
Also, as two different images cannot be mapped into the same latent features, the bijective metric is more reliable.
It should be noticed that the direct access to $p_I(\mathbf{x})$ in Eqn. \ref{eqn:GeneratorFormulation} is unavailable since we do not know the dataset set used to learn the mapping function $F$. 
However, the approximation can be efficiently and practically adopted according to the prior knowledge about $p_I(\mathbf{x})$ that are facial images. 
Let $p_x(\mathbf{x})$ be a density function, an approximation of $p_I(\mathbf{x})$, estimated from an alternative large-scale face dataset.
Therefore, the optimization of $G$ in Eqn. \eqref{eqn:GeneratorFormulation_Dist}, with an approximation $p_x(\mathbf{x})$ by drawing images from a large-scale face dataset, can be rewritten as:
\begin{equation} \label{eqn:GeneratorFormulation_Dist}
\begin{split}
    \theta_g &\approx \arg \min_{\theta} \int \mathcal{L}_G^x(\tilde{\mathbf{x}},\mathbf{x}) p_x(\mathbf{x}) d\mathbf{x}
\end{split}
\end{equation}
where $\tilde{\mathbf{x}} = [G \circ F](\mathbf{x}; \theta)$.
Now, let $H: \mathcal{I} \mapsto \mathcal{Z}$ be a bijective function that maps $\mathbf{x}$ in the image space to a latent variable $\mathbf{z}=H(\mathbf{x})$ in the deep latent space.
With the bijective property, the optimization in Eqn. \eqref{eqn:GeneratorFormulation_Dist} is equivalent to. 
\begin{equation} \label{eqn:GeneratorFormulation_Bijection}
\begin{split}
    &\arg \min_{\theta} \int \mathcal{L}_G^z(H(\tilde{\mathbf{x}}),H(\mathbf{x})) p_x(\mathbf{x}) d\mathbf{x}\\
    =& \arg \min_{\theta} \int \mathcal{L}_G^z(H(\tilde{\mathbf{x}}),H(\mathbf{x})) p_z(\mathbf{z})|\det(\mathbf{J_x^{\top}} \mathbf{J_x})|^{1/2} d\mathbf{z}\\
    =& \arg \min_{\theta} \int \mathcal{L}_G^z(\tilde{\mathbf{z}},\mathbf{z}) p_z(\mathbf{z})|\det(\mathbf{J_x^{\top}} \mathbf{J_x})|^{1/2} d\mathbf{z}\\
\end{split}
\end{equation}
where $\tilde{\mathbf{z}}=H(\tilde{\mathbf{x}})$; $p_x(\mathbf{x})=p_z(\mathbf{z})|\det(\mathbf{J_x^{\top}} \mathbf{J_x})|^{1/2}$ by the change of variable formula; $\mathbf{J_x}$ is the Jacobian of $H$ with respect to $\mathbf{x}$; and $\mathcal{L}_G^z$ is the distance metric in $\mathcal{Z}$. 
Intuitively, Eqn. \eqref{eqn:GeneratorFormulation_Bijection} indicates that instead of computing the distance $\mathcal{L}_G^x$ and estimating $p_x(\mathbf{x})$ directly in the image domain, the optimization process can be equivalently accomplished via the distance $\mathcal{L}_G^z$ and density $p_z(\mathbf{z})$ in $\mathcal{Z}$ according to the bijective property of $H$.

\subsection{The Generalizability of Bijective Metric}

In the previous section, we intuitively illustrated the equivalence of the optimization of the Bijective Metric between image space and the deep latent space in Eqn. \eqref{eqn:GeneratorFormulation_Bijection}. This section will prove the generalizability of the Bijective Metric over the standard image reconstruction loss.
Without the lack of generality, we assume that $\mathcal{L}_G^x$ is the Euclidean distance commonly used for image reconstruction.

\noindent
\textbf{Proposition 1:} \textit{Given a bijective function  $H: \mathcal{I} \mapsto \mathcal{Z}$, the direct metric $\mathcal{L}_G^x$ is bounded by the Bijective Metric $\mathcal{L}_G^z$}
\begin{equation}
    \mathcal{L}_G^x(\tilde{\mathbf{x}}, \mathbf{x}) = \Theta\left(\mathcal{L}_G^z(\tilde{\mathbf{z}}, \mathbf{z})\right) \quad \text{where} \quad \tilde{\mathbf{z}} = H(\tilde{\mathbf{x}}), \mathbf{z} = H(\mathbf{x})
\end{equation}

\noindent
\textbf{Proof:}
Since the mapping function $H$ is a bijective mapping, it has been proved that the bijective function is locally Lipschitz continuous \cite{pmlr-v189-verine23a}:
\begin{equation} \label{eqn:lip_1}
    \forall \mathbf{x}, \tilde{\mathbf{x}} \in \mathcal{X}: \quad ||H(\tilde{\mathbf{x}}) - H(\mathbf{x})||_2 \leq L||\tilde{\mathbf{x}} - \mathbf{x}||_2
\end{equation}
where $L$ is the Lipschitz constant. In addition, due to the bijective property of $H$, the inverse function of $H$ is also locally Lipschitz continuous \cite{pmlr-v189-verine23a}:
\begin{equation} \label{eqn:lip_2}
    \forall \mathbf{z}, \tilde{\mathbf{z}} \in \mathcal{Z}: \quad ||H^{-1}(\tilde{\mathbf{z}}) - H^{-1}(\mathbf{z})||_2 \leq L'||\tilde{\mathbf{z}} - \mathbf{z}||_2
\end{equation}
The details of the Lipschitz continuity of the bijective network have been proved in \cite{pmlr-v189-verine23a}.
From Eqn. \eqref{eqn:lip_1} and Eqn. \eqref{eqn:lip_2}, the Lipschitz continuity of the bijective mapping $H$ can be expressed over $H$ as follows:
\begin{equation} \label{eqn:final_proof}
\small
\begin{split}
    \forall \mathbf{x}, \tilde{\mathbf{x}} \in \mathcal{X}: \quad 
    & \quad \frac{1}{L'}||\tilde{\mathbf{x}} - \mathbf{x}||_2 \leq ||H(\tilde{\mathbf{x}}) - H(\mathbf{x})||_2 \leq L||\tilde{\mathbf{x}} - \mathbf{x}||_2  \\
    \Leftrightarrow& \quad
     \frac{1}{L'}\mathcal{L}_G^x(\tilde{\mathbf{x}}, \mathbf{x}) \quad  \leq \quad \mathcal{L}_G^z(\tilde{\mathbf{z}}, \mathbf{z}) \quad \leq \quad  L\mathcal{L}_G^x(\tilde{\mathbf{x}}, \mathbf{x}) \\
     \Rightarrow& \quad \mathcal{L}_G^x(\tilde{\mathbf{x}}, \mathbf{x}) = \Theta\left(\mathcal{L}_G^z(\tilde{\mathbf{z}}, \mathbf{z})\right) 
\end{split}
\end{equation}

Formally, under this asymptotic bound in Eqn. \eqref{eqn:final_proof}, by optimizing our Bijective Metric clustering loss, the quantitative reconstruction loss on the image space $\mathcal{L}_G^x(\tilde{\mathbf{x}}, \mathbf{x})$ has also been explicitly imposed. 
Going beyond the property of generalizability stated in \textbf{Proposition 1}, our approach offers other benefits over direct metrics. 
In particular, our approach is able to provide high-quality image reconstruction while maintaining the reconstruction of features of the same subject identification being similar thanks to the deep latent representations produced by the bijective network.

\subsection{The Prior Distributions $p_z$.}
In general, there are various choices for the prior distribution $p_z$ and the ideal one should have two properties: (1) \textit{simplicity in density estimation}, and (2) \textit{easily sampling}. 
Motivated by these properties, Gaussian distribution has been chosen for $p_z$. 
It should be noticed that any other distribution types are still applicable in our framework if the chosen distribution satisfies the mentioned properties.

\subsection{The Large-Margin Contrastive Distance Metric} %

Due to the choice of $p_z$ as a Gaussian distribution, 
the distance between images in $\mathcal{I}$ is equivalent to the deviation between Gaussians in the deep latent space. 
Therefore, we can effectively define $\mathcal{L}_G^z$  as the squared Wasserstein coupling distance between two Gaussian distributions.
\begin{equation} \label{eqn:GaussDistance}
\begin{split}
    \mathcal{L}_G^z(\tilde{\mathbf{z}},\mathbf{z})
    =& d(\tilde{\mathbf{z}},\mathbf{z})
    = \inf \mathbb{E}(||\mathbf{\tilde{z}}-\mathbf{z}||_2^2)\\
    =& || \tilde{\mu} - \mu ||^2_2 
    + \text{Tr}(\tilde{\Sigma} + \Sigma - 2(\tilde{\Sigma}^{1/2}\Sigma\tilde{\Sigma}^{1/2})^{1/2})
\end{split}
\end{equation}
where $\{\tilde{\mu}, \tilde{\Sigma}\}$ and $\{\mu, \Sigma\}$ are the means and covariances of $\mathbf{\tilde{z}}$ and $\mathbf{z}$, respectively.
The metric $\mathcal{L}_G^z$ then can be further extended with image labels to improve the contrastive learning capability by reducing the distance between images of the same identity and enhancing the margin between different identities.
\begin{equation}
    \mathcal{L}_G^{z_{id}}(\mathbf{\tilde{z}_1},\mathbf{\tilde{z}_2}) = 
    \begin{cases}
    d(\mathbf{\tilde{z}_1},\mathbf{\tilde{z}_2})       &  \text{if } l_{\mathbf{\tilde{z}_1}} = l_{\mathbf{\tilde{z}_2}}\\
    \max(0,d(\mathbf{\tilde{z}_1},\mathbf{\tilde{z}_2}) - m)  & \text{if } l_{\mathbf{\tilde{z}_1}} \neq l_{\mathbf{\tilde{z}_2}}
  \end{cases}
\end{equation}
where $m$ denotes the hyper-parameter controlling the margin between classes; 
and  $\{l_{\mathbf{\tilde{z}_1}}, l_{\mathbf{\tilde{z}_2}}\}$ denote the subject identity of $\{\mathbf{\tilde{z}_1}, \mathbf{\tilde{z}_2}\}$, respectively.

\subsection{Learning the Bijection}
To efficiently learn the bijective mapping $H$, the multi-scale architecture network with tractable log Jacobian determinant computation \cite{dinh2016density, Duong_2017_ICCV} has been adopted as the backbone network.
Additionally, to further improve the discriminative capability in the of $H$ in the deep latent space $\mathcal{Z}$, the identity labels have been exploited during the training process of $H$. In particular, given $N$ classes of the training set, the Gaussian distributions with different means $\{\mu_i\}_{i=1}^N$ and covariances $\{\Sigma_i\}_{i=1}^N$ have been employed to enforce samples of each class distributed to its own prior distribution, i.e., $\mathbf{z}_k \sim \mathcal{N}(\mu_k, \Sigma_k)$. 
Formally, the mapping $H$ can be learned by the negative log-likelihood loss formulated as follows:
\begin{equation}
\small
\begin{split}
    \theta_H^* &= -\arg \min_{\theta_H} \mathbb{E}_{\mathbf{x} \sim p_x(\mathbf{x})}\log p_x(\mathbf{x}, k;\theta_H) \\
    &= -\arg \min_{\theta_H} \mathbb{E}_{\mathbf{x} \sim p_x(\mathbf{x})} \left[\log p_z(\mathbf{z}, k;\theta_H) + \frac{1}{2}\log |\det(\mathbf{J_x^{\top}} \mathbf{J_x})|\right] %
\end{split}
\end{equation}

\subsubsection{Structuring of Bijection Latent Space with ID labels}
Different from previous works where only a Gaussian distribution, i.e., Normal Distribution, is designed for the Bijection's Latent space, multiple Gaussians are required to form the Latent space where images of each subject ID/class distribute within a Gaussian distribution in that latent space. 
In order to improve the discriminative property of $H$ in latent space $\mathcal{Z}$, we propose to exploit the ID label in the training process of $H$.
Particularly, the means and covariances of Gaussian distributions are pre-defined for all $C$ classes:
\begin{equation}
\begin{split}
    \boldsymbol{\mu}_c &= \text{\textbf{1}}c; \boldsymbol{\Sigma} = \mathbf{I}\\ 
    \mathbf{z}_c &\sim \mathcal{N}(\boldsymbol{\mu}_c, \mathbf{I})
\end{split}
\end{equation}
where \text{\textbf{1}} is the all-one vector, the subscript $c$ denotes for class $c$.

\subsection{Learning from Distillation Knowledge} \label{sec:Distillation}

In the traditional approach, the generator $G$ can still be learned by adopting the perceptual distance \cite{liu2021infinite} to compute the difference between deep feature representations, i.e., $F(\tilde{\mathbf{I}})$ and $F(\mathbf{I})$. However, as aforementioned in Section \ref{sec:LimitationClassifierMetric}, the direct access to the network $F$, i.e., the gradients and the intermediate representations, is unavailable due to the blackbox setting. 
To maximize the information that can be exploited from the recognition engine $F$, 
we propose to distill the matching knowledge from the blackbox $F$ to a ``student`` function $F^S$. Then, distilled knowledge adapted to $F^S$ will be employed to train the generator $G$.
There are two important purposes shown by the distillation process. First, the student network $F^S$ can mimic the behavior of the network $F$ by aligning its deep feature space to the one produced by $F$ and maintaining the semantic information of extracted features in the reconstruction process. Second, with our designed $F^S$, the knowledge about the embedding process of $F$, i.e., the network gradients and intermediate representations, will become transparent. Therefore, knowledge from the blackbox function $F$ can be maximally exploited via the student function $F^S$.

Mathematically, let $F^S: \mathcal{I} \mapsto \mathcal{F}$ and $F^S=F^S_1 \circ F^S_2 \cdots \circ F^S_n$ be the composition of $n$-sub components. The knowledge from $F$ can be distilled to $F^S$ by aligning their extracted features defined as follows:
\begin{equation} \label{eqn:LearningStudentMatcher}
\begin{split}
    \theta_S =& \arg \min_{\theta_S} \mathcal{L}_S = \mathbb{E}_{\mathbf{x} \sim p_x} d_{distill}\left(F(\mathbf{x}),F_S(\mathbf{x}; \theta_S) \right)\\
    =& \arg \min_{\theta_S} \mathbb{E}_{\mathbf{x} \sim p_x} \left\| 1 - \frac{F(\mathbf{x})}{\parallel F(\mathbf{x})\parallel}* \frac{F^S(\mathbf{x}; \theta_S)}{\parallel F^S(\mathbf{x}; \theta_S)\parallel}\right\|^2_2
\end{split}
\end{equation}
Hence, the generator $G$ can be further enhanced via the distilled knowledge of not only final embedding features but also the intermediate representations via the student function $F^S$ formed as follows:
\begin{equation} \label{eqn:DistillationLoss}
\begin{split}
    \mathcal{L}_G^{distill}(\mathbf{\tilde{x}}, \mathbf{x}) &= \sum_{j=1}^n \lambda_j \frac{\left\| F_j^S(\mathbf{\tilde{x}};\theta_S) - F_j^S(\mathbf{x};\theta_S) \right\|}{W_j H_j C_j} \\
    +& \lambda_a \left\| 1 - \frac{F^S(\mathbf{\tilde{x}};\theta_S)}{\parallel F^S(\mathbf{\tilde{x}};\theta_S)\parallel}* \frac{F^S(\mathbf{x}; \theta_S)}{\parallel F^S(\mathbf{x}; \theta_S)\parallel}\right\|^2_2
\end{split}
\end{equation}
where $\{\lambda_j\}^n_1$ and $\lambda_a$ denote the hyper-parameters controlling the impacts of two terms. 
The first term of $\mathcal{L}_G^{distill}(\mathbf{\tilde{x}}, \mathbf{x})$ aims to penalize the differences between the intermediate structure of the desired and reconstructed facial images while the second term validates the similarity of their final deep features.

\section{Attention-based Bijective Generative Adversarial Networks} \label{sec:GeneratorLearning}

In this section, we will present the training procedure of our proposed network.
Figure \ref{fig:ProposedFramework} illustrates the proposed framework with Bijective Metric and Distillation Process mentioned in the previous sections to train the generator $G$.

\subsection{Training Procedure}

Given the input image $\mathbf{x}$, the generator $G$ takes the deep feature $F(\mathbf{x})$ as the input and aims to reconstruct the image $\tilde{\mathbf{x}}$ so that the reconstructed image $\tilde{\mathbf{x}}$  is as similar to the original image $\mathbf{x}$ as possible in terms of identity and visual appearance. The GAN-based generator architecture \cite{danihelka2017comparison} is adopted for our generator structure $G$. The entire framework is optimized by using different criteria defined as follows.
\begin{equation}
\begin{split}
    \mathcal{L}_G =& \lambda_{b} \mathcal{L}^{biject} + \lambda_d \mathcal{L}^{distill} + \lambda_{adv} \mathcal{L}^{adv} + \lambda_{r} \mathcal{L}^{recon}\\
   \mathcal{L}^{biject} =& \mathbb{E}_{\mathbf{x}\sim p_x} \left[ \mathcal{L}_G^x \left([G \circ F](\mathbf{x}; \theta), \mathbf{x}\right)\right]\\
   +& \mathbb{E}_{\mathbf{x}_1,\mathbf{x}_2\sim p_x} \left[ \mathcal{L}_G^{x_{id}} \left([G \circ F](\mathbf{x}_1; \theta), [G \circ F](\mathbf{x}_2;\theta\right)\right]\\
   =&\mathbb{E}_{\mathbf{z}\sim p_z} \left[ \mathcal{L}_G^z \left(\mathbf{\tilde{z}}, \mathbf{z}\right)\right] + \mathbb{E}_{\mathbf{z}_1,\mathbf{z}_2\sim p_z} \left[ \mathcal{L}_G^{z_{id}} \left(\mathbf{\tilde{z}}_1, \mathbf{\tilde{z}}_2\right)\right]\\
   \mathcal{L}^{distill} =& \mathbb{E}_{\mathbf{x}\sim p_x} \left[ \mathcal{L}_G^{distill} \left([G \circ F](\mathbf{x}; \theta), \mathbf{x}\right)\right]\\
   \mathcal{L}^{adv} =& \mathbb{E}_{\mathbf{x}\sim p_x} \left[ D\big([G \circ F](\mathbf{x};\theta)\big)\right]\\
   \mathcal{L}^{recon} =& \mathbb{E}_{\mathbf{x}\sim p_x} \left[ \big\| [G \circ F](\mathbf{x}) - \mathbf{x} \big\|_1\right]
\end{split}
\end{equation}
where $\{\mathcal{L}^{biject},\mathcal{L}^{distill}, \mathcal{L}^{adv}, \mathcal{L}^{recon}\}$ denote the bijective, distillation, adversarial, and reconstruction losses, respectively. $D$ is a discriminator distinguishing the real images from synthesized ones. $\{\lambda_{b}, \lambda_d, \lambda_{adv}, \lambda_{r}\}$ are their parameters controlling their relative importance.
The bijective metric $\mathcal{L}^{biject}$ and the distillation loss $\mathcal{L}^{distill}$ constrain the identity preservation.
The reconstruction loss $\mathcal{L}^{recon}$ enforces the similarity between the original image and the reconstructed image; meanwhile, via the discriminator $D$, the adversarial loss $\mathcal{L}^{adv}$ penalizes the realism of the synthesized image. The discriminator $D$ is alternately updated along with the generator $G$ as follows.
\begin{equation}
\begin{split}
    \mathcal{L}_D =& \mathbb{E}_{\mathbf{x} \sim p_{x}} \left[ D\big([G \circ F](\mathbf{x};\theta)\big) \right] - \mathbb{E}_{\mathbf{x} \sim p_{x}} \left[ D\big(\mathbf{x})\big) \right]\\
    +& \lambda \mathbb{E}_{\mathbf{\hat{x}} \sim p_{\hat{x}}} \left[ \left( \|\nabla_{\mathbf{\hat{x}}}D(\mathbf{\hat{x}})\|_2 -1 \right)^2\right]
\end{split}
\end{equation}
where $p_{\hat{x}}$ is the random interpolation distribution between real and generated images \cite{gulrajani2017improved}.
Finally, the GAN-based minimax strategy is adopted to train the entire framework.

\subsection{Exponential Weighting Decay}

As the progressive growing training manner \cite{karras2017progressive} starts learning from the low resolution and gradually increasing the level of details, it is efficient for learning the images in general. Nevertheless, the strategy is inefficient in terms of maintaining the identity information. Particularly, in the initial scales at low resolutions where the face images are still blurry, it is difficult to control the subject identity of faces to be synthesized. Meanwhile, in the later scales when the generator becomes more mature and learns to add more details, the identities of those faces have already been structured and become hard to change. There is a trade-off between realism and identity preservation during the progressive training process. Therefore, we propose an exponential weighting scheme for emphasizing identity preservation in the early stages and penalizing realism in the later stages. Formally, the hyper-parameter set $\{\lambda_{b}, \lambda_d, \lambda_{adv}, \lambda_{r}\}$ is progressively defined as follows:
\begin{equation}
\begin{split}
    \lambda_{b}   &= \alpha e^{R_M - R(i)} \\ 
    \lambda_{d}   &= e^{R_M - R(i)} \\
    \lambda_{adv} &= \beta e^{R(i)} \\ 
    \lambda_{r}   &= e^{R_M - R(i)} 
\end{split}
\end{equation}
where $R(i)$ denotes the current scales of stage $i$ and $R_M$ is the maximum scales to be learned by $G$.

\section{Experiments} \label{sec:Experiments}

In this section, we first describe the datasets and metrics, followed by the model configurations used in our experiments. Then, we present our ablative experiments to illustrate the effectiveness of our proposed approach from different perspectives. Finally, we compared our experimental results with other approaches to illustrate the state-of-the-art performance of our proposed approach.

\subsection{Datasets and Metrics}

\noindent
\textbf{Datasets:}
The training dataset includes the publicly available Casia-WebFace \cite{yi2014learning} consisting of 490K labeled facial images of over 10K subjects.
Since we focus on the open-set scenario, the duplicated subjects between training and testing sets are removed to ensure no overlapping between them.
For validation, as common practices of attribute learning and image quality evaluation, 
we adopt the testing split of 10K images from CelebA \cite{liu2015faceattributes} to validate the reconstruction quality. 
For ID preservation, we explore 
{LFW \cite{huang2008labeled}, CFP-FP \cite{sengupta2016frontal}, CP-LFW \cite{CPLFWTech}, AgeDB \cite{moschoglou2017agedb}, CA-LFW \cite{DBLP:journals/corr/abs-1708-08197}}
which provide the standard face verification protocols against different in-the-wild face variations.
Since each face matcher engine requires a different preprocessing process, the training and testing data are aligned to the required template accordingly.

\noindent
\textbf{Metrics:}
To evaluate the quality of the reconstructed images, we adopt two metrics, i.e., Inception Score \cite{salimans2016improved} and Multi-Scale Structural Similarity \cite{odena2017conditional}. The Inception Score is a common metric used to assess the quality of synthetic images. The score is calculated based on the output produced by the pre-trained Inceptionv3 \cite{szegedy2016rethinking} model. Meanwhile, the Multi-Scale Structural Similarity metric is a generalized form of the Structural Similarity Index Measure that quantifies image quality degradation. Multi-Scale Structural Similarity is conducted over multiple scales through a process of multiple stages of sub-sampling.
To evaluate the ability of our model in terms of ID preservation, we adopt the face verification metric \cite{nguyen2010cosine}. 
In particular, given a pair of images, we randomly synthesize either one of two images from its feature extracted from the corresponding matcher.
Then, the face verification metric is performed based on the concise distance between the facial features of two image queries.

\subsection{Model Configurations}

\noindent
\textbf{Network Architectures. }
For our CNN-based generator, the generator structure from PO-GAN \cite{karras2017progressive} is utilized which consists of 5 convolutional blocks for $G$ whereas the feature conditional branch has 8 fully connected layers. The combination of five consecutive blocks of two convolutions and one downsampling operator is included in the discriminator $D$. The adoption of minibatch-stddev operator continued by convolution and fully connected is done in the last block of $D$. The application of AdaIN operator \cite{huang2017arbitrary} is done for the feature injection node. A configuration is set up for the bijection $H$, which has 5 sub-mapping functions where each of them is represented as two 32-feature-map residual blocks. 
The training of this structure is done using the log-likelihood objective function on Casia-WebFace. Resnet-50 \cite{he2016deep} is adopted for $F^S$.

Meanwhile, the structure from TransGAN \cite{jiang2021transgan} is adopted for our attention-based generator. Particularly, the attention-based transformer has 5 blocks of Transformer. Instead of using the vanilla self-attention layer in each block of Transformer, the memory-efficient grid self-attention is utilized to reduce the memory bottleneck when the resolution is scaling up to high resolution.
In addition, the facial feature is progressively injected into the generator by the cross-attention mechanism \cite{vaswani2017attention}.
The multi-scale discriminator from \cite{jiang2021transgan} is adopted for our discriminator $D$. In our paper, we denote the CNN-based generator and attention-based generator as DiBiGAN and DAB-GAN, respectively.

\noindent
\textbf{Optimization Configurations. }
The entire framework is implemented in Tensorflow and all the models are trained on a machine with four GPUs of NVIDIA RTX Quadro P8000.
The batch size is set based on the resolution of output images. For the very first resolution of output images ($4 \times 4$), the batch size is set to $128$. Then, the batch size will be divided by two when the resolution of the images is doubled.
The Adam Optimizer with the started learning rate of $0.0015$  has been employed to optimize the network.
We empirically set $\{\alpha = 0.001, \beta = 1.0, \lambda_j=1, \lambda_a = 10.0\}$ to maintain the balanced values between loss terms.

\subsection{Ablation Study}

This section first studies the impact of the Bijective Metrics. Then, we analyze the effectiveness of our method in the reconstructed results in different aspects, including (1) Frontal Facial Features, (2) Occlusions and Expressions, (3) Background Information, and (4) Different Features of the Same Subjects. 

\subsubsection{Effectiveness of Bijective Metrics}

\begin{figure}[!t]
	\centering \includegraphics[width=0.98\columnwidth]{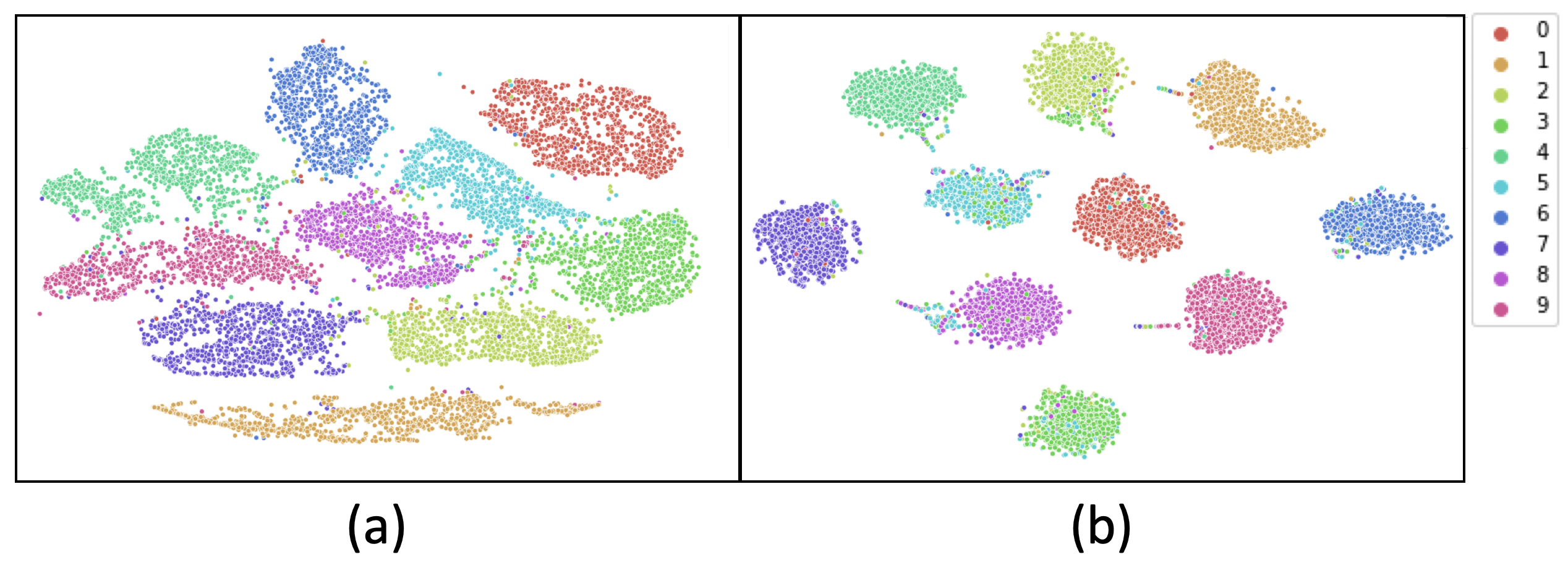}
	\caption{\textbf{The distributions of synthesized MNIST samples} on the testing set (a) without, and (b) with adopting Bijective Metric. 
	}%
	\label{fig:MNISTDistribution}
\end{figure}

To illustrate the effectiveness of our proposed bijective metric for the feature reconstruction task, we conduct an ablative experiment on MNIST \cite{lecun1998mnist} with LeNet \cite{lecun1998gradient} as the function $F$.
To remove the effects of other factors, we consider the whitebox mode in this experiment where $F$ is directly used in $\mathcal{L}^{distill}$. 
The entire training set of MNIST and their $1 \times 1024$ feature vectors are used to train the generator $G$. Since the image size of MNIST is $32\times 32$, we only use three convolution blocks for our CNN-based generator and discriminator.
The resulting distributions of synthetic testing images of all classes without and with our proposed bijective metric $\mathcal{L}^{biject}$ are illustrated in Fig. \ref{fig:MNISTDistribution}. 
As shown in this figure, the distribution with our bijective metric learning (Fig. \ref{fig:MNISTDistribution}(b)) is supervised with a more direct metric learning mechanism in the image domain and, therefore, shows the advantages with enhanced intra-class and inter-class distributions compared to the case where the generator $G$ is learned with only classifier-based metrics (Fig. \ref{fig:MNISTDistribution}(a)).

\begin{figure}[t]
	\centering \includegraphics[width=1.0\columnwidth]{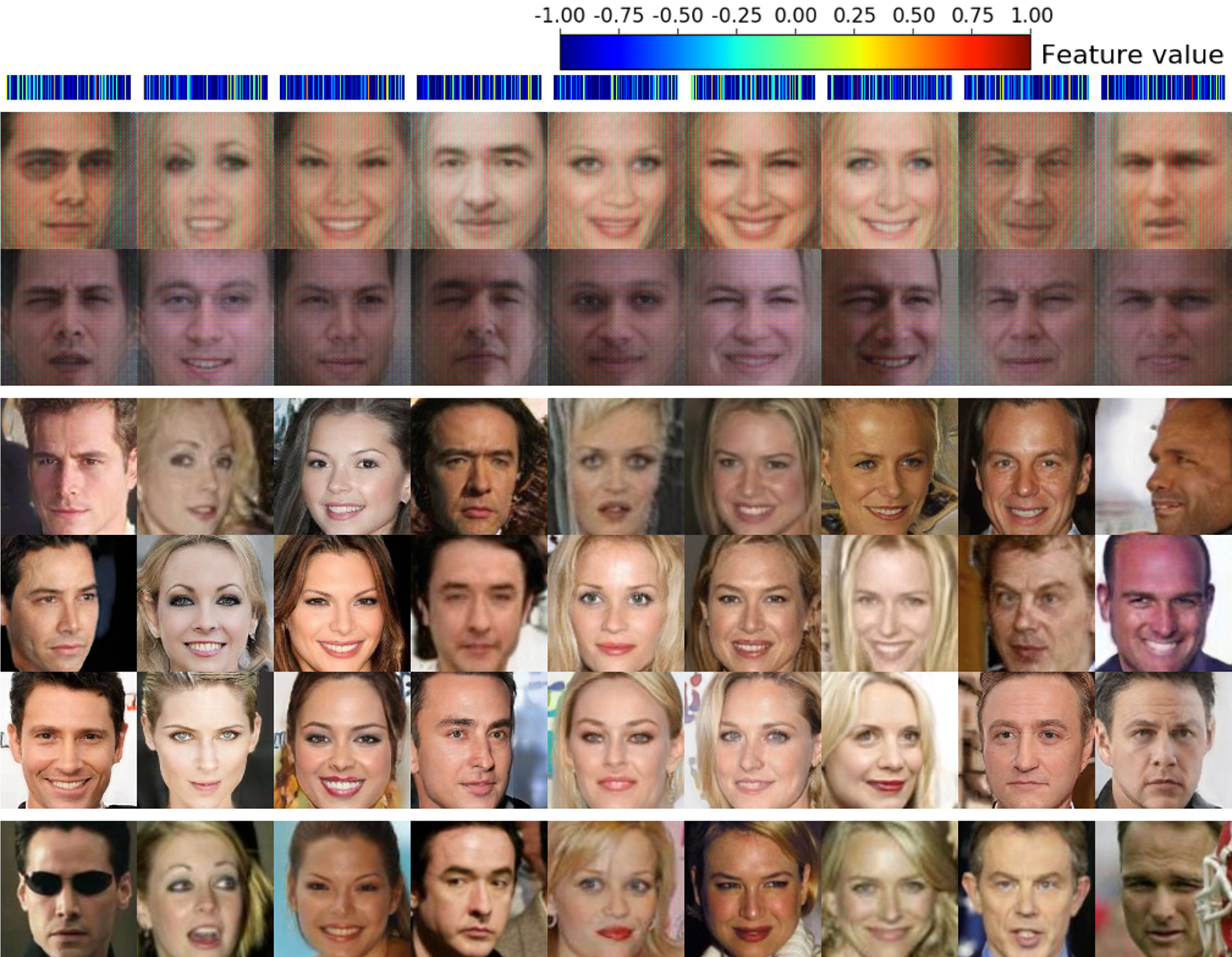}
	\caption{\textbf{Feature Reconstruction against in-the-wild facial variations.} For each subject, given an input feature (1st row), while  VGG-NBNet \cite{mai2018reconstruction} and MPIE-NBNet \cite{mai2018reconstruction} (2nd and 3rd rows) reconstruct faces with limited quality, DiBiGAN in whitebox mode (4th row), DiBiGAN in blackbox mode (5th row), and DAB-GAN in blackbox mode (6th row)  are able to produce realistic faces with better ID preservation comparable to real faces (6th row).} 
	\label{fig:LFWRecon}
\end{figure}

\begin{figure*}[t]
	\centering \includegraphics[width=1.0\textwidth]{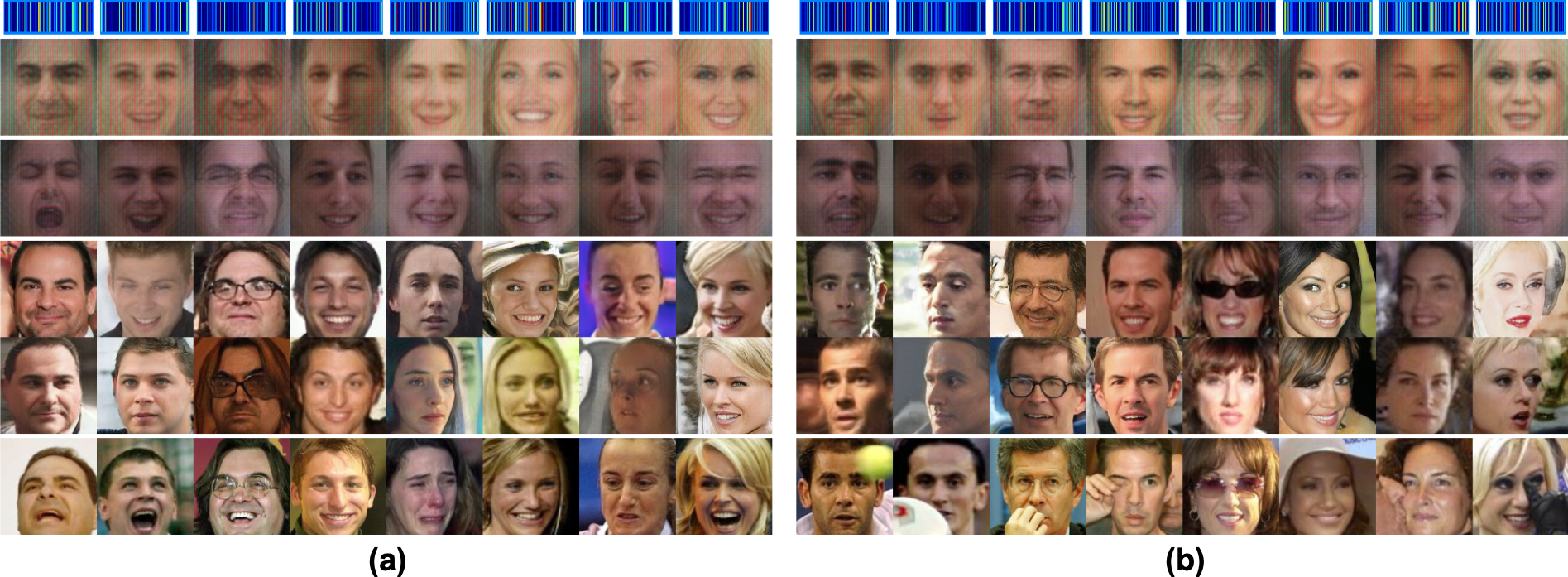}
	\caption{\textbf{Feature Reconstruction against expressions (a) and occlusions (b).} For each subject, the 1st row shows the input feature. The next five rows are VGG-NBNet \cite{mai2018reconstruction}, MPIE-NBNet \cite{mai2018reconstruction}, Our DibiGAN in whitebox and blackbox settings, and Real Faces, respectively.}
	\label{fig:ReconExpressionOcclusion}
\end{figure*}

\subsubsection{Effectiveness of Frontal Facial Features}

With our proposed method,
given only the deep features extracted from $F$ on testing images, the generator $G$ is able to 
to reconstruct the subjects' faces.
Fig. \ref{fig:LFWRecon} illustrates the qualitative comparison of our synthetic faces with other baselines.
In particular, our generator $G$ is able to reconstruct realistic faces even when their embedding features are extracted from faces with a wide range of in-the-wild variations. 
Importantly, our proposed method successfully preserves the ID features of these subjects in both whitebox and blackbox settings. 
In the whitebox setting, since the structure of $F$ is accessible, the learning process can effectively exploit different aspects of the embedding process from $F$ and produce a generator $G$ that depicts better facial features of the real faces. 
Meanwhile, although the accessible information is very limited in blackbox setting, the learned generator $G$ is still able to observe the distilled knowledge of $F^S$ and efficiently fill the knowledge gap in the whitebox setting. 
As shown in Fig. \ref{fig:LFWRecon}, we achieve better face in terms of image quality and ID preservation compared to NBNet \cite{mai2018reconstruction}. Moreover, the quality of synthetic faces reconstructed by the attention-based generator is better than the ones synthesized by the CNN-based generator.

\subsubsection{Effectiveness of Occlusions and Expressions}

Fig. \ref{fig:ReconExpressionOcclusion} demonstrates our synthesis based on attributes of faces with both expressions and occlusions. Similarly to the previous experiment, our model portrays realistic faces with comparable ID features as in real faces.
The quality of the reconstructed faces consistently outperforms NBNet in both realistic and ID categories.
The success of robustly dealing with those challenging factors can be attributed to two factors: (1) the matcher $F$ was trained to ignore those facial variations in its embedding features; and (2) both bijective metric learning and distillation processes can efficiently exploit necessary knowledge from $F$ as well as real face distributions in image domain for the synthesis process.

\subsubsection{Effectiveness of Background Information}
\begin{figure}[!b]
	\centering \includegraphics[width=1.0\columnwidth]{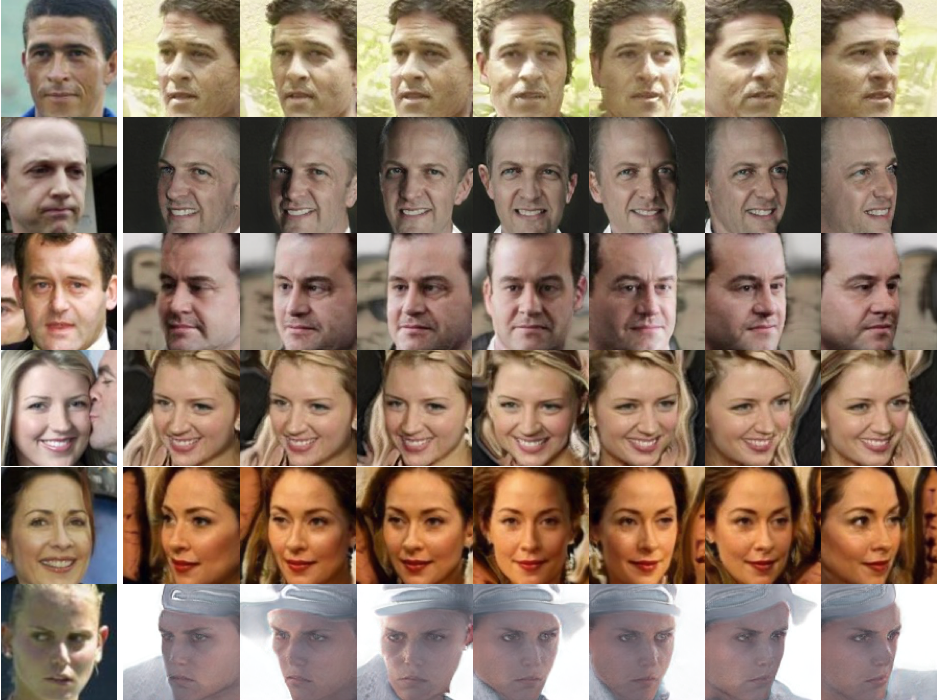}
	\caption{From the input features, our model can synthesize various conditions of a face by varying the ``background'' variable $\mathbf{v}$.}
	\label{fig:ReconDifferentPoses}
\end{figure}

As mentioned in Sec. \ref{sec:Generator}, the variable $\mathbf{v}$ is utilized to represent background factors, allowing $G$ to focus on modeling ID characteristics. As a result, multiple situations of that face may be synthesized by fixing the input feature and modifying the variable values, as illustrated in Fig. \ref{fig:ReconDifferentPoses}. These findings highlight the benefits of our model structure in terms of its capacity to capture numerous aspects of the reconstruction process.

\begin{figure}[!t]
	\centering \includegraphics[width=1.0\columnwidth]{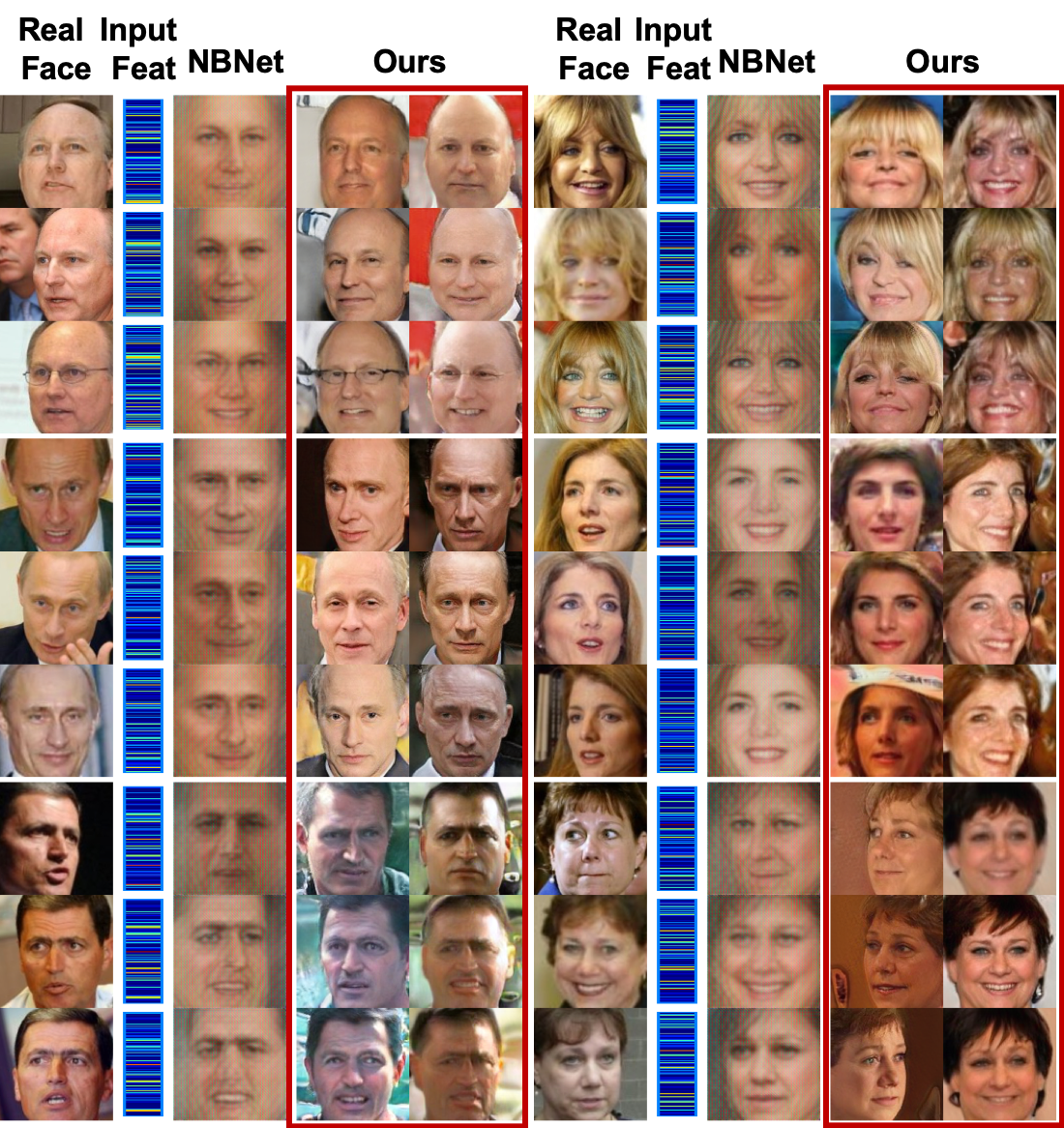}
	\caption{\textbf{Feature Reconstruction against features of the same subject.} For each subject, the first and second columns show different real faces and their features of a subject. Compared to VGG-NBNet \cite{mai2018reconstruction} (third column), our DibiGAN in whitebox and blackbox modes can effectively preserve the ID of the subjects.}
	\label{fig:ReconSameSubject}
\end{figure}

\input{tables/whitebox-blackbox}

\subsubsection{Effectiveness of Different Features of Same Subject}

Fig. \ref{fig:ReconSameSubject} demonstrates the benefits of our technique for synthesizing faces from diverse feature representations of the same subject. These findings highlight the benefits of the proposed bijective metric in improving the border between classes and constraining the similarity of reconstructed faces of the same subject in the image domain. As a consequence, reconstructed faces from the same subject ID not only retain the features of that subject (i.e., are comparable to actual faces) but also share similar features.

\subsection{Comparison with State-of-the-Art Methods}

In this section, we compare our reconstructed results without prior state-of-the-art methods in two different criteria, i.e., the face verification performance and the image quality. Our experimental results have shown that our proposed method has achieved state-of-the-art performance in terms of preserving facial identity and image quality compared to prior methods.

\subsubsection{Face Verification Results}

In order to quantitatively evaluate the realism of our synthetic faces and how well the proposed approach can preserve the ID feature, three metrics are adopted: (1) Face Verification Accuracy \cite{nguyen2010cosine}; (2) Multi-scale Structural similarity (MS-SSIM) \cite{odena2017conditional}; (3) Inception Score (IS) \cite{salimans2016improved};

\noindent
\textbf{ID Preservation.} Our model is tested against LFW, CFP-FP, CP-LFW, AgeDB, and CA-LFW with an image in each positive pair replaced by the reconstructed one and the remaining image of that pair preserved as the reference actual face. Table \ref{tab:MatchingAccuracyAll} reports the matching accuracy. Each component in our framework highlights the advantages and contributions through these results.
In comparison to the PO\_GAN structure, our Feature-Conditional Structure (DiBiGAN) allows for greater flexibility in modeling ID characteristics and achieves higher matching accuracy. In conjunction with distilled information from $F^S$, the Generator delivers a significant increase in accuracy, closing the gap to actual faces to only 2.02\% and 2.722\% on LFW in whitebox and blackbox settings, respectively. 
By including the bijective measure, these gaps are decreased to 0.6\% and 0.65\% for the two settings, respectively. 
Moreover, with the attention-based generator (DAB-GAN), we further improve the verification performance and reduce the gap between synthetic faces and real faces in both whitebox and blackbox settings. 

\noindent
\textbf{Reconstructions against different Face Recognition Engines.} To demonstrate the effectiveness of our proposed approach, we tested it against several face recognition engines, as shown in Table \ref{table:BlackboxDifferentMatcher}. All generators are set to blackbox mode, which means that only the final extracted features are available. Our reconstructed faces can retain the ID information and attain the same level of accuracy as actual faces.
These results highlight our model's effectiveness in capturing the behaviors of the feature extraction functions $F$ and providing high-quality reconstructions and maintaining the verification performance.

\subsubsection{Face Quality Results}

To evaluate the realism of the synthetic faces, we synthesize all faces of CelebA testing set in several training configurations as shown in Table \ref{tab:MatchingAccuracyAll} where each loss function in cumulatively enables on the top of the previous configuration.
As the baseline protocol \cite{karras2017progressive}, we compared MS-SSIM and IS metrics between our synthetic faces and other baselines (PO\_GAN \cite{karras2017progressive} and NBNet \cite{mai2018reconstruction}) in both whitebox and blackbox settings. 
It should be noted that the adversarial and reconstruction losses for configs (A), (D), and (G).
For all configs (A), (B), and (C), the PO\_GAN baseline takes only the embedding features as its input.
As shown in Table \ref{tab:MatchingAccuracyAll}, the qualitative results have shown that our approach maintains the competitive reconstruction quality as PO\_GAN and has a small gap between our synthetic faces and  real faces.
Moreover, the quantitative results of our synthetic images consistently outperform NBNet in both MS-SSIM and IS metrics.

\input{tables/different_matchers}

\section{Limitations and Broader Impacts}

\noindent
\textbf{Limitations:} Although our method can reconstruct the facial images that preserve the identity of the subjects, the size of our synthesized images remains low resolution. This limitation is caused by the training data. In particular, the current face recognition models often use low-resolution data (i.e., $112 \times 112$) in their training and testing. This work follows the standard protocol of face recognition to develop our method to reconstruct the facial images that match with the face recognition protocol. As a matter of fact, our proposed method can totally scale up the resolution of the reconstructed output with respect to the high-resolution training data.

\noindent
\textbf{Broader Impacts:}
This work introduces a novel bijective metric learning for the feature reconstruction task, i.e., reconstructing the facial images from the feature vector. The reconstructed faces can be used for improper purposes, i.e., attack the face recognition engines. In addition, the reconstructed faces could reveal sensitive information about users in the face recognition system, including appearance, age, gender, ethnicity, or other factors.
Our work does not intend to attack the face recognition system. In fact, we aim to develop a novel bijective metric learning for the feature reconstruction task and study the effectiveness of the generator network and the distillation process in the image reconstruction problem.
Also, throughout our work, we have raised the vulnerability of the current deep learning-based face recognition models. This line of research will open a new direction in studying safe and protected face recognition systems.

\section{Conclusions} %
\label{sec:Conclusion}

This work has introduced a novel attention-based structure with Bijective Metric Learning for the feature reconstruction problem to reconstruct the subjects' faces given their deep blackboxed features.
The feature reconstruction problem is considered in the context of blackbox setting and open-set problem which makes the problem becoming challenging.
Thanks to the proposed Bijective Metric, the presented attention-based generator, and the Distillation Knowledege, our DAB-GAN effectively maximizes the information to be exploited from a given blackbox face recognition engine.
In addition, the generalizability of the Bijective Metrics has been further clarified and proved based on the Lipschitz continuity property.
The intensive experiments on a wide range of in-the-wild face variations against different face-matching engines have demonstrated the advantages of our method in synthesizing realistic faces with the subject's identity preservation.

\bibliographystyle{IEEEtran}
\bibliography{references}

\end{document}

%% file: tables/whitebox-blackbox.tex
\begin{table*}[b] 
    \centering
    \caption{\textbf{Realism Quality and Matching Accuracy.} Comparison results in Multi-Scale Structural Similarity (MSSIM) (\textit{the smaller value is better}); Inception Score (IS) and Matching Accuracy (\textit{the higher value is better}). For each configuration in (A)-(C), (D)-(F), and (G)-(I), each loss function is cumulative and enabled on top of the previous configuration. 
    }
    \setlength{\tabcolsep}{2pt}
    \label{tab:MatchingAccuracyAll}
    \resizebox{\textwidth}{!}{  
        \begin{tabular}{|l|cc|ccccc|cc|ccccc|}
        \toprule
\multicolumn{1}{|c|}{}          & \multicolumn{7}{c|}{\textbf{Whitebox   Reconstruction}}                                                                                                           & \multicolumn{7}{c|}{\textbf{Blackbox Reconstruction}}                                                                                                             \\
\cline{2-15}
                              & \multicolumn{2}{c|}{\textbf{Celeb A}} & \multirow{2}{*}{\textbf{LFW}} & \multirow{2}{*}{\textbf{CFP-FP}} & \multirow{2}{*}{\textbf{CP-LFW}} & \multirow{2}{*}{\textbf{AgeDB}} & \multirow{2}{*}{\textbf{CA-LFW}} & \multicolumn{2}{c|}{\textbf{Celeb A}} & \multirow{2}{*}{\textbf{LFW}} & \multirow{2}{*}{\textbf{CFP-FP}} & \multirow{2}{*}{\textbf{CP-LFW}} & \multirow{2}{*}{\textbf{AgeDB}} & \multirow{2}{*}{\textbf{\textbf{CA-LFW}}} \\
\multicolumn{1}{|c|}{}          & \textbf{MSSIM}       & \textbf{IS}          &                      &                         &                        &                        &                        & \textbf{MSSIM}       & \textbf{IS}          &                      &                         &                        &                        &                        \\
\midrule
Real Face                     & 0.305         & 3.008       & 99.78\%              & 97.10\%                 & 92.08\%                & 98.40\%                & 95.45\%               
& 0.305         & 3.008       & 99.70\%              & 93.10\%                 & 90.12\%                & 96.80\%                & 93.82\%                \\
\midrule
VGG-NBNet \cite{mai2018reconstruction}                     & $-$             & $-$           & $-$                    & $-$                       & $-$                      & $-$                      & $-$                      & 0.661         & 1.387       & 91.42\%              & 74.63\%                 & 72.10\%                & 80.42\%                & 73.65\%                \\
MPIE-NBNet \cite{mai2018reconstruction}                    & $-$             & $-$           & $-$                    & $-$                       & $-$                      & $-$                      & $-$                      & 0.592         & 1.484       & 93.17\%              & 78.51\%                 & 74.18\%                & 79.45\%                & 71.93\%                \\
\midrule
(A) PO\_GAN                    & \textbf{0.331}         & \textbf{2.226}       & 68.20\%              & 68.89\%                 & 64.99\%                & 63.42\%                & 67.17\%                & \textbf{0.315}         & 2.227       & 66.63\%              & 65.59\%                 & 65.62\%                & 62.37\%                & 66.26\%                \\
(B) + $\mathcal{L}^{distill}$ & 0.343         & 2.073       & 96.03\%              & 79.07\%                 & 78.28\%                & 83.33\%                & 79.06\%                & 0.337         & \textbf{2.238}       & 94.95\%              & 78.80\%                 & 77.74\%                & 81.56\%                & 78.94\%                \\
(C) + $\mathcal{L}^{biject}$  & 0.358         & 2.052       & \textbf{98.10\%}              & \textbf{88.01\%}                 & \textbf{86.51\%}                & \textbf{88.16\%}                & \textbf{84.33\%}                & 0.360         & 2.176       & \textbf{97.30\%}              & \textbf{82.51\% }                & \textbf{85.86\%}                & \textbf{85.71\%}                & \textbf{82.77\%}                \\
\hline
(D) DiBiGAN                   & 0.316         & 2.343       & 79.82\%              & 81.71\%                 & 74.49\%                & 77.20\%                & 75.49\%                & 0.305         & \textbf{2.463}       & 77.57\%              & 80.66\%                 & 73.27\%                & 76.83\%                & 74.80\%                \\
(E) + $\mathcal{L}^{distill}$ & \textbf{0.306}         & 2.349       & 97.76\%              & 89.20\%                 & 86.19\%                & 86.61\%                & 84.57\%                & 0.305         & 2.423       & 97.06\%              & 84.83\%                 & 85.76\%                & 91.70\%                & 83.04\%                \\
(F) + $\mathcal{L}^{biject}$   & 0.310         & \textbf{2.531}       & \textbf{99.18\%}              & \textbf{92.67\%}                 & \textbf{88.30\%}                & \textbf{94.18\%}                & \textbf{91.40\%}                & \textbf{0.303}         & 2.422       & \textbf{99.13\%}              & \textbf{89.03\%}                 & \textbf{87.26\%}                & \textbf{93.53\%}                & \textbf{89.44\%}                \\
\midrule
(G) DAB-GAN                   & 0.315         & 2.361       & 82.41\%              & 82.62\%                 & 77.31\%                & 80.12\%                & 79.37\%                & 0.321         & 2.425       & 79.28\%              & 81.85\%                 & 76.20\%                & 78.72\%                & 78.18\%                \\
(H) + $\mathcal{L}^{distill}$ & \textbf{0.302}         & \textbf{2.425}       & 99.02\%              & 92.17\%                 & 82.84\%                & 92.25\%                & 85.87\%                & 0.318         & \textbf{2.442}       & 98.96\%              & 86.72\%                 & 80.56\%                & 92.35\%                & 84.89\%                \\
(I) + $\mathcal{L}^{biject}$  & 0.319         & 2.367       & \textbf{99.52\%}              & \textbf{93.11\%}                 & \textbf{89.36\% }               & \textbf{95.58\%}                & \textbf{92.73\%}                & \textbf{0.301}         & 2.322       & \textbf{99.21\%}              & \textbf{92.17\%}                 & \textbf{88.27\%}                & \textbf{94.18\%}                & \textbf{90.38\%}               \\
\bottomrule
\end{tabular}
    }
\end{table*}

%% file: tables/different_matchers.tex
\begin{table}[!t]
\centering
\caption{Accuracy against different blackbox face matchers.}\label{table:BlackboxDifferentMatcher}
\setlength{\tabcolsep}{2pt}
\resizebox{\columnwidth}{!}{  
\begin{tabular}{|c|c|ccccc|}
\toprule
\textbf{Matcher}                         & \textbf{Approach} & \textbf{LFW}     &\textbf{ CFP-FP}  & \textbf{CP-LFW}   & \textbf{AgeDB}   & \textbf{CA-LFW}   \\
\midrule
\multirow{3}{*}{ArcFace \cite{deng2019arcface}}        & Real     & 99.78\% & 97.10\% & 92.08\% & 98.40\% & 95.45\% \\
                                & DiBiGAN  & 99.13\% & 89.03\% & 87.26\% & 93.53\% & 89.44\% \\
                                & \textbf{DAB-GAN}  & \textbf{99.21\%} & \textbf{92.17\%} & \textbf{88.27\%} & \textbf{94.18\%} & \textbf{90.38\%} \\
\midrule
\multirow{3}{*}{FaceNet \cite{schroff2015facenet}}        & Real     & 99.55\% & 94.05\% & 92.59\% & 90.16\% & 95.26\% \\
                                & DiBiGAN  & 98.05\% & 87.19\% & 82.04\% & \textbf{89.80\%} & 88.64\% \\
                                & \textbf{DAB-GAN}  & \textbf{98.25\%} & \textbf{88.51\%} & 85.59\% & \textbf{89.80\%} & \textbf{90.07\%} \\
\midrule
\multirow{3}{*}{SphereFacePlus \cite{LiuNIPS18}} & Real     & 98.92\% & 91.16\% & 92.14\% & 91.92\% & 95.98\% \\
                                & DiBiGAN  & 97.21\% & 86.86\% & 81.08\% & 88.98\% & 87.94\% \\
                                & \textbf{DAB-GAN}  & \textbf{97.95\%} & \textbf{87.75\%} & \textbf{83.26\%} & \textbf{89.12\%} & \textbf{90.19\%} \\
\midrule
\multirow{3}{*}{MagFace \cite{meng2021magface}}        & Real     & 99.83\% & 98.46\% & 92.87\% & 98.17\% & 96.15\% \\
                                & DiBiGAN  & 97.48\% & 92.56\% & 83.21\% & 92.84\% & 91.14\% \\
                                & \textbf{DAB-GAN}  & \textbf{98.65\%} & \textbf{94.78\%} & \textbf{86.16\%} & \textbf{95.53\%} & \textbf{93.27\%} \\
\midrule
\multirow{3}{*}{SCF-ArcFace \cite{li2021spherical}}    & Real     & 99.82\% & 98.40\% & 93.16\% & 98.30\% & 96.12\% \\
                                & DiBiGAN  & 97.38\% & 91.12\% & 82.55\% & 93.22\% & 91.18\% \\
                                & \textbf{DAB-GAN}  & \textbf{98.58\%} & \textbf{93.85\%} & \textbf{86.54\%} & \textbf{95.72\%} & \textbf{94.98\%} \\
\midrule
\multirow{3}{*}{AdaFace \cite{kim2022adaface}}        & Real     & 99.80\% & 99.17\% & 94.63\% & 97.90\% & 96.05\% \\
                                & DiBiGAN  & 96.88\% & 95.18\% & 83.56\% & 86.01\% & 88.47\% \\
                                & \textbf{DAB-GAN}  & \textbf{98.60\% }& \textbf{97.62\%} & \textbf{88.41\%} & \textbf{91.54\%} & \textbf{92.04\%} \\
\bottomrule
\end{tabular}
}
\end{table}